\documentclass[lettersize,journal]{IEEEtran}
\usepackage{amsmath,amsfonts}
\usepackage{algorithm}
\usepackage{algpseudocode}
\usepackage{array}
\usepackage[caption=false,font=normalsize,labelfont=sf,textfont=sf]{subfig}
\usepackage{textcomp}
\usepackage{stfloats}
\usepackage{url}
\usepackage{verbatim}
\usepackage{graphicx}
\usepackage{cite}
\usepackage{float}
\usepackage[pagebackref,breaklinks,colorlinks]{hyperref}
\usepackage{booktabs}
\usepackage{multirow}
\usepackage[table]{xcolor}
\usepackage{colortbl}
\usepackage{pifont}
\newcommand{\cmark}{\ding{52}}%
\newcommand{\xmark}{\ding{56}}%
\definecolor{tableHeadGray}{gray}{.9}

\newcommand{\Add}[1]{\textcolor{black}{#1}}

\begin{document}

\title{Event-Based Motion Estimation via Oriented Distance Fields}

\author{Lei~Sun$^{*}$, Yuqin~Ma$^{*}$, Weilun~Li, Haoran~Liang, Runyi~Yang, Kaiwei~Wang, Danda~Pani~Paudel, and~Luc~Van~Gool
\thanks{$^{*}$L. Sun and Y. Ma contributed equally to this work.}
\thanks{L. Sun, H. Liang, R. Yang, D. P. Paudel, and L. Van Gool are with INSAIT, Sofia University ``St. Kliment Ohridski'', Sofia, Bulgaria.}
\thanks{Y. Ma, W. Li, and K. Wang are with Zhejiang University, Hangzhou, China.}}

\markboth{}%
{Sun \MakeLowercase{\textit{et al.}}: Event-Based Motion Estimation via Oriented Distance Fields}


\maketitle

\begin{abstract}
\Add{Event-based motion estimation is central to tasks that demand high temporal resolution and robustness to fast motion. Existing methods typically rely on iterative optimization or repeated hypothesis comparison, offsetting the sensor's low-latency advantage. We propose Oriented Distance Field Motion Estimation (ODF Motion Estimation), which replaces this optimization with a single averaging step over a precomputed field of event distance vectors, combined with an adaptive event-count selection strategy and a parameter-free trail filter. On public and self-collected datasets, ODF motion estimation reaches sub-pixel accuracy at the lowest latency among compared methods. We validate its generality on two downstream applications rather than treating them as separate contributions. First, the estimated trajectory is converted into a blur kernel and paired with a compact iterative-unfolding network, trained on simulated motion-estimation noise, for real-time non-blind image deblurring, attaining competitive or superior PSNR/SSIM with under 1M parameters. Second, the same precomputed field is repurposed for directional event filtering in a low-power asynchronous pupil and glint tracker, sustaining stable tracking for tens of seconds while lowering a near-eye module's power draw.}
\end{abstract}

\begin{IEEEkeywords}
\Add{Event cameras, motion estimation, image deblurring, eye tracking.}
\end{IEEEkeywords}

\section{Introduction}
\label{sec:intro}

\Add{\IEEEPARstart{R}{ecovering} the relative motion between a camera and its scene within a short observation window is a fundamental perception capability for platforms that must operate while moving quickly, mobile robots equipped with gimbal or Pan-Tilt-Zoom (PTZ) camera systems chief among them, which rely on rapid panning and tilting for wide-area situational awareness. An event camera is well matched to this need: rather than fixed-rate frames, it reports per-pixel brightness changes asynchronously and with microsecond resolution~\cite{gallego2020event}, so the stream itself directly encodes the motion trajectory rather than requiring it to be inferred after the fact. Recovering this trajectory means aligning the edges implied by early events against those implied by each subsequent event, and existing solutions fall into two families: an explicit optimization objective solved iteratively~\cite{gallego2018unifying,kim2021real,nunes2021robust,huang2023progressive}, or a discrete search over hypothesized states compared per incoming event~\cite{alzugaray:3DV19,alzugaray:BMVC20}.}

\Add{Both families reintroduce the very latency that motivates using an event camera in the first place. Iterative optimization requires repeatedly evaluating and refining an energy function before it converges, and hypothesis-based search requires comparing several candidate states for every event rather than resolving it directly. For a platform whose onboard perception must keep pace with its own fast angular motion, this per-event computational burden, not the sensor's raw temporal resolution, becomes the binding constraint on how quickly motion can actually be recovered.}

\Add{This paper removes that bottleneck by replacing optimization or search with a single closed-form operation. We propose Oriented Distance Field Motion Estimation (ODF motion estimation), which precomputes, once per edge template, a field whose value and descent direction at every location directly encode the translation that would align that location with the template. So estimating a new event's motion reduces to a lookup into this field, and combining many events reduces to averaging those lookups, with no per-event optimization, search, or learned inference. Under the common case of a dominant 2-DoF translation, stated explicitly and examined further in Sec.~\ref{sec:motion_estimation}, this closed-form average recovers a trajectory competitive with what iterative or search-based solvers need many more operations to reach.}

\begin{figure*}[!htbp]
    \centering
    \includegraphics[width=1.\textwidth]{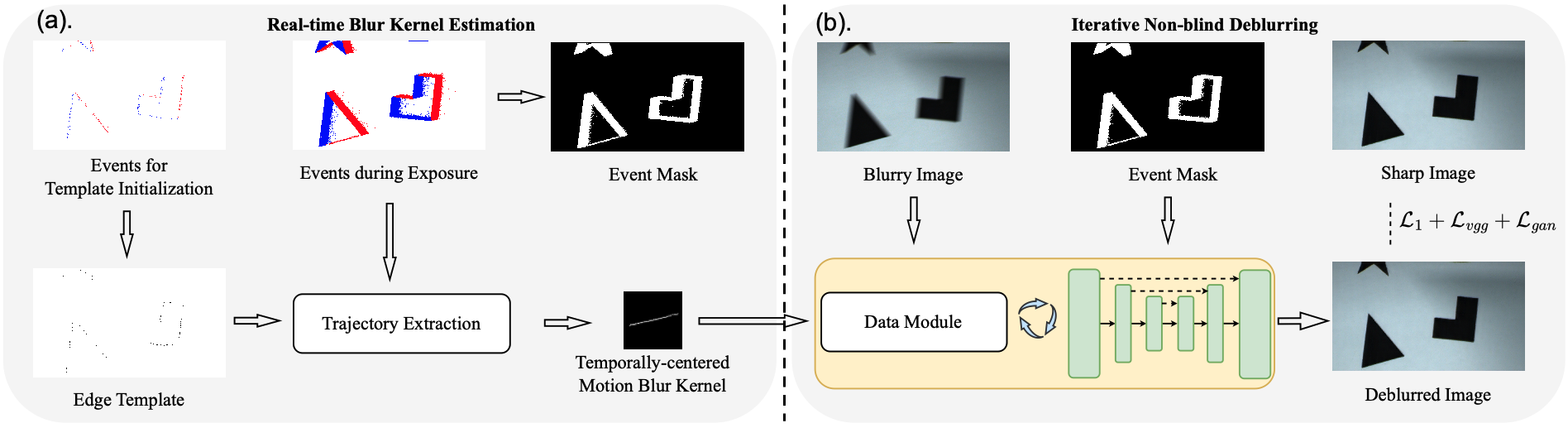}
    \caption{\Add{Overview of the proposed pipeline. (a) ODF motion estimation (Sec.~\ref{sec:motion_estimation}). An edge template built from events or a reference frame is used to precompute a distance field and its descent direction. Averaging the distance vectors of subsequent events against this field gives the motion trajectory during exposure, which yields a temporally-centered blur kernel and an event mask marking where events occurred. (b) Image deblurring with the estimated kernel (Sec.~\ref{sec:deblurring}). The kernel and event mask from (a), together with the blurry image, are passed to an iterative deep-unfolding network that alternates a closed-form data module with a learned denoising network to recover the sharp image. The same motion trajectory from (a) is also reused as a directional filter for asynchronous pupil and glint tracking (Sec.~\ref{sec:eye_tracking}).}}
    \label{fig:pipeline}
    \vspace{-17pt}
\end{figure*}

\Add{Because the trajectory ODF motion estimation recovers is useful wherever motion itself must be known, not specific to one task, we validate it through two applications that consume it in different ways rather than presenting them as independent contributions. The trajectory is converted into a blur kernel to test whether the estimate is accurate enough to support non-blind image deblurring, and the same underlying field is repurposed as a directional event filter to test whether the estimator generalizes to asynchronous feature tracking, demonstrated on pupil and glint tracking for eye tracking. Both serve to validate the estimator under different accuracy and robustness requirements rather than to propose new deblurring or eye-tracking architectures in their own right (Fig.~\ref{fig:pipeline}).}

\Add{In summary, we deliver the following contributions:}
\begin{itemize}
    \item \Add{ODF motion estimation, a closed-form algorithm for event-based 2-DoF motion estimation that replaces iterative optimization or hypothesis search with a single distance-vector average, an adaptive event-count selection strategy, and a parameter-free trail filter.}
    \item \Add{Validation of ODF motion estimation's generality through two downstream consumers of its output, a non-blind deblurring network and a directional event filter for asynchronous tracking, evaluated as tests of the estimator rather than as separate contributions.}
    \item \Add{Extensive evaluation on public and self-collected datasets, showing that ODF motion estimation reaches state-of-the-art or competitive accuracy at substantially lower latency than optimization- or search-based alternatives, and that this accuracy carries through to competitive downstream deblurring and tracking performance at a fraction of the parameter count.}
\end{itemize}

\section{Related Work}
\label{sec:related_work}

\subsection{Motion Estimation with Event Cameras}
\Add{Event-based motion estimation methods either repurpose classical optical flow machinery~\cite{liu2018adaptive,almatrafi2020distance} or exploit the event generation model directly, as in contrast maximization~\cite{gallego2018unifying}, time-surface alignment~\cite{huang2023progressive}, and closed-form line solvers~\cite{gao2024n}. Multi-hypothesis trackers~\cite{alzugaray:BMVC20} compare many candidate states per event. Frame-fusion methods such as EKLT~\cite{gehrig2020eklt} and edge-aligned tracking~\cite{kuse2016robust} extend feature lifetime through iterative alignment. ODF motion estimation instead precomputes the distance vector field once per template, reducing motion estimation to a single average per event.}

\subsection{Blurry Kernel Estimation and Deblurring}
Image deblurring methods split by whether the blur kernel is known: non-blind deblurring recovers the sharp image from a known or estimated kernel via classical solvers or learned priors such as USRNet~\cite{zhang2020deep}, DPIR~\cite{zhang2022plugandplay}, and DWDN~\cite{dong2021dwdn}, while blind deblurring estimates both the kernel and the sharp image from the blurred input alone. \Add{Event cameras give non-blind deblurring a direct route to the kernel: the event double integral model~\cite{edi_pan,lin2023fast} recovers it in closed form but is noise-sensitive, while end-to-end fusion networks~\cite{sun2022event,sun2024unified} skip kernel estimation entirely yet generalize poorly beyond their training event statistics~\cite{shen2024restoring}. ODF motion estimation keeps the pipeline compact and non-blind, closing this gap with an explicit motion-estimation-noise kernel model rather than iterative optimization or heavy fusion.}

\subsection{Eye Tracking}
\Add{Eye tracking most commonly locates the pupil via pupil-center-corneal-reflection (PCCR)~\cite{guestrin2006general} or appearance-based models, both requiring fast, stable pupil (and, for PCCR, glint) detection. Event cameras avoid the power-accuracy trade-off of a fixed frame rate, and event eye tracking has progressed from asynchronous model fitting~\cite{angelopoulos2021event} to learned pupil segmentation and gaze regression~\cite{zhao2023eveye,li2024egaze}, with spiking implementations on neuromorphic hardware~\cite{bonazzi2024retina} reaching milliwatt-level power but still trailing frame-based PCCR in accuracy. We build on this line of work by reusing the translation vectors from Sec.~\ref{sec:motion_estimation} for directional event filtering rather than learning a separate pupil model, keeping the tracker parameter-free.}

\section{Event-Based Motion Estimation}
\label{sec:motion_estimation}

\Add{ODF motion estimation relies on three conditions holding over the events used for a single trajectory estimate. The scene motion is globally consistent, so one 2-DoF translation explains every tracked edge rather than independently moving regions. That translation is well approximated as 2-DoF rather than a general rotation, affine, or projective warp. And the template's edges span enough distinct orientations to constrain both components of the translation. These conditions are what make a closed-form average a substitute for iterative optimization, and Sec.~\ref{sec:motion_estimation} returns to each: dynamic-scene failure for the first, specific camera platforms for the second, and Sec.~\ref{subsec:field_generate} for the third, where it sets the accuracy of the averaged estimate. Violating a condition degrades the estimate in a specific way rather than uniformly: an independently moving object biases the average toward a mixture of motions, unmodeled rotation or higher-order warp aliases into the estimate as systematic error, and insufficient orientation diversity biases the under-constrained component toward the dominant edges' direction, consequences we revisit in our limitations (Sec.~\ref{sec:conclusion}).}

\subsection{Edge Alignment}

Motion events are triggered predominantly near scene edges, where the spatial gradient is large, since the intensity increment during motion is
\begin{equation}
    \Delta L \approx - \nabla L \cdot \mathbf{v} \Delta t,
\end{equation}
where $L$ is the (log) image intensity, $\mathbf{v}$ is the instantaneous velocity of the scene relative to the camera, and $\Delta t$ is a small time interval. The resulting event stream therefore records these edges' trajectory during exposure, which we recover through edge alignment: matching the edges formed by the earliest events against those formed by each subsequent event. Edge or feature alignment has been applied between frame images~\cite{kuse2016robust}, point clouds, and event streams~\cite{zhou2021event,kim2021real,nunes2021robust,huang2023progressive}. We apply the same principle between an edge template and incoming events.

\begin{figure}[!t]
    \centering
    \begin{minipage}{0.16\textwidth}\centering
    \includegraphics[width=\textwidth]{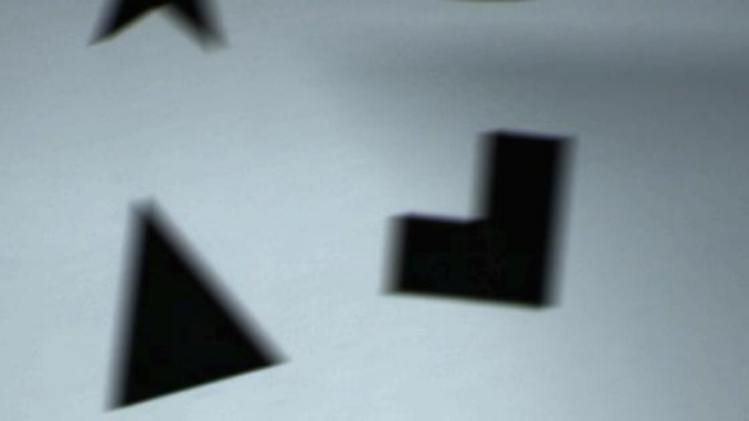}\\[2pt]{\scriptsize (a)}
    \end{minipage}\hspace{1mm}
    \begin{minipage}{0.16\textwidth}\centering
    \includegraphics[width=\textwidth]{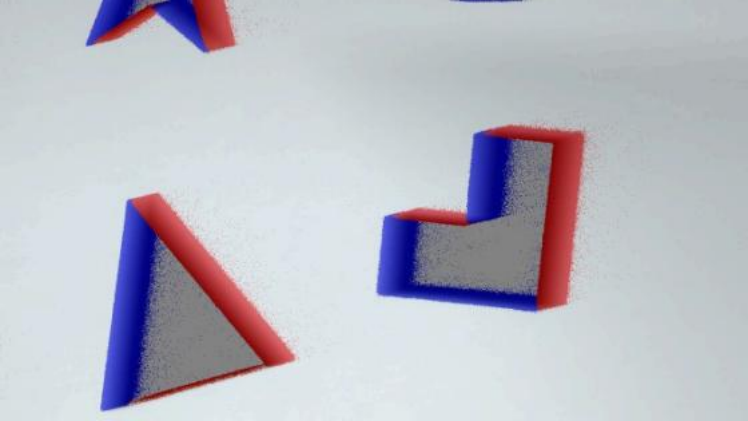}\\[2pt]{\scriptsize (b)}
    \end{minipage}\hspace{1mm}
    \begin{minipage}{0.137\textwidth}\centering
    \includegraphics[width=\textwidth]{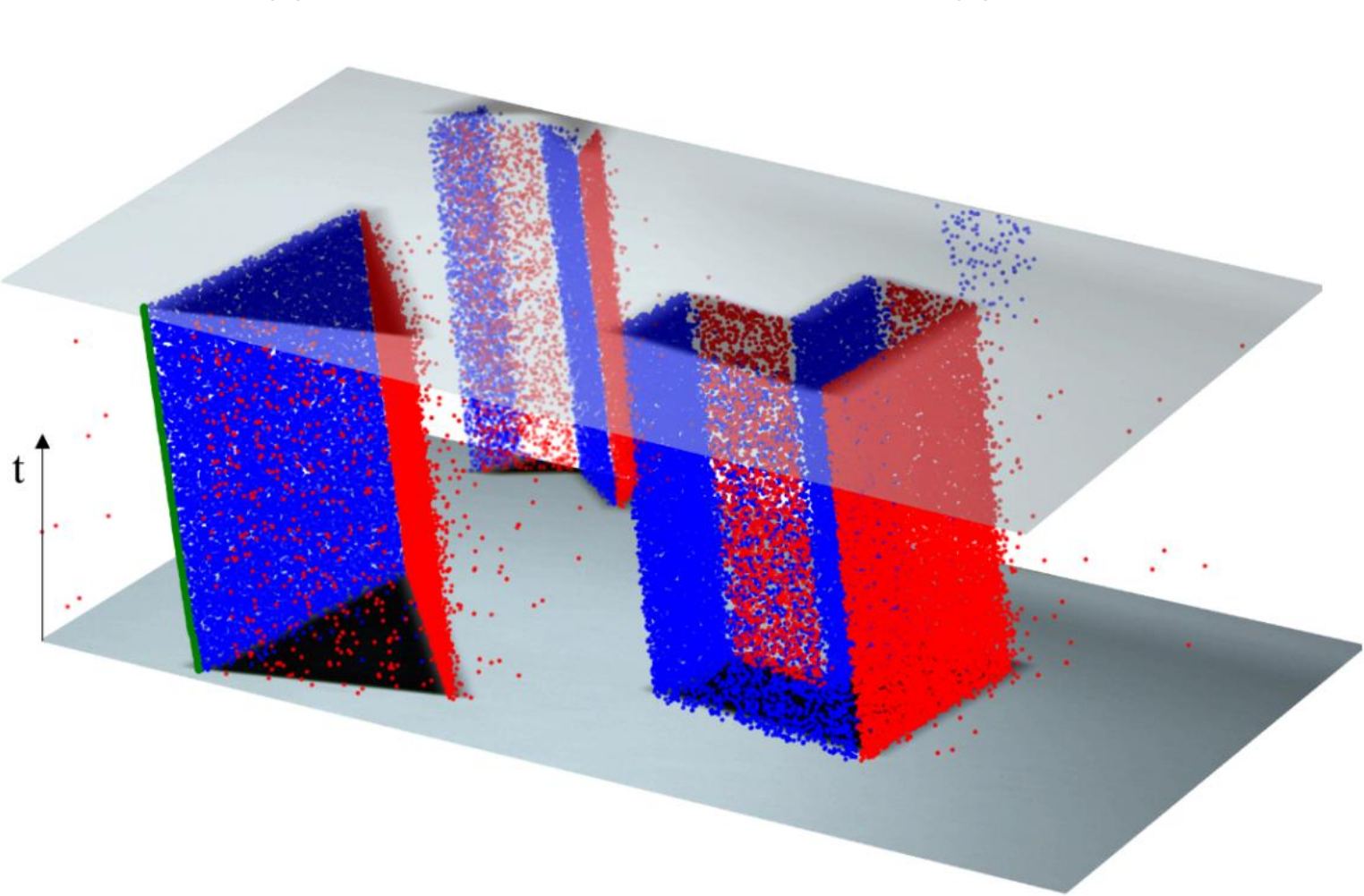}\\[2pt]{\scriptsize (c)}
    \end{minipage}
    \caption{Event-based trajectory extraction. (a) Blurred frame. (b) Events within the exposure window. (c) Event stream (red and blue dots) with the extracted motion trajectory (green line).}
    \label{fig:blur_and_event}
    \vspace{-17pt}
\end{figure}

Blur induced by 2-degree-of-freedom (2-DoF) translation, such as handshake or panning, is the common case we target. Existing event-based edge tracking methods address this and more general motion via an optimization objective solved through iterative~\cite{gallego2018unifying,kim2021real,nunes2021robust,huang2023progressive} or discrete search~\cite{alzugaray:3DV19,alzugaray:BMVC20} procedures. We show this optimization can be replaced entirely under the 2-DoF assumption: trajectory recovery reduces to exploiting event temporal continuity together with an oriented distance field, without any iterative or discrete search step (Fig.~\ref{fig:blur_and_event}).

\subsection{Pattern Generation}
\label{subsec:pattern_generation}

\begin{algorithm}
\caption{Pattern Generation With Adaptive Batch Strategy}
\label{alg:adaptive}
\begin{algorithmic}[1]
\State \textbf{Input:} events $E$, down sampling factor $N$, overlap ratio $r_i$
\State create event count frame $I_{count}$ to record the overlap
\State slice $E$ into event bundles {$\xi_1, \xi_2, ...$}, each containing $n_e$ events
\For{$\xi_i$ in {$\xi_1, \xi_2, ...$}}
    \State $n_{overlap} \gets 0$ \Comment{reset for the current bundle}
    \For{each event $(x, y, t, p) \in \xi_i$}
        \State $(x', y') \gets (x // N, y // N)$
        \If{$I_{count}(x', y') \neq 0$}
            \State $n_{overlap} \gets n_{overlap} + 1$
        \EndIf
        \State $I_{count}(x', y') \gets I_{count}(x', y') + 1$

    \EndFor
    \If{$n_{overlap} / n_e > r_i$}
        \State collect the preceding events {$\xi_1, ..., \xi_i$}
        \State break
    \EndIf
\EndFor

\end{algorithmic}
\end{algorithm}

Building an edge template from a fixed time window or event count is unreliable: edge density and motion speed vary across scenes, so a fixed threshold yields templates too sparse to constrain motion or too dense to compute efficiently. Prior adaptive schemes~\cite{huang2023progressive, liu2018adaptive} address this but add their own overhead. \Add{We instead exploit an adjacency effect already present in spatial down sampling: as events are down sampled, nearby events increasingly collide with already-occupied bins, and the fraction of colliding events within each newly processed bundle signals when the template has accumulated sufficient spatial coverage. We track this fraction per bundle, resetting it before each new bundle, and stop once it exceeds the ratio $r_i$ (Algorithm~\ref{alg:adaptive}), set empirically rather than from a coverage model since collision rate depends on scene content in ways we do not model explicitly. The choice is not delicate: once a template nears sufficient coverage, further events collide with occupied bins at a fast-rising rate, so small changes to $r_i$ shift the stopping point by a few events rather than which templates are accepted.}

\subsection{Oriented Distance Field}
\label{subsec:field_generate}

\Add{Given the edge template of Sec.~\ref{subsec:pattern_generation}, we construct a field over it that encodes, at every pixel, the translation needed to align that pixel with the reference edge.} Distance fields built from events have proven effective for optical flow estimation~\cite{almatrafi2020distance} and related motion tasks. Following the use of time surfaces as distance fields~\cite{zhou2021event}, we build a field from the events nearest the start of exposure, align each subsequent event against it by shifting to minimize distance, and take the accumulated shift over all processed events as the edge trajectory.

\Add{The underlying idea is geometric: if the edge has moved by translation $t$ since the reference template was built, every point on the moved edge sits exactly $t$ away from its corresponding reference point, so a field storing this direction and distance at every location implicitly stores a translation estimate everywhere at once, precomputed rather than solved per event.}

A standard distance field measures alignment quality but does not by itself yield the warp parameters needed to align two edges. Under the 2-DoF assumption, we instead construct an oriented distance field \Add{$D: \Omega \to \mathbb{R}_{\geq 0}$ over the image domain $\Omega$} that directly encodes the translation parameters at every location.

\begin{figure}[!t]
    \centering
    \begin{minipage}{0.22\textwidth}\centering
    \includegraphics[width=\textwidth]{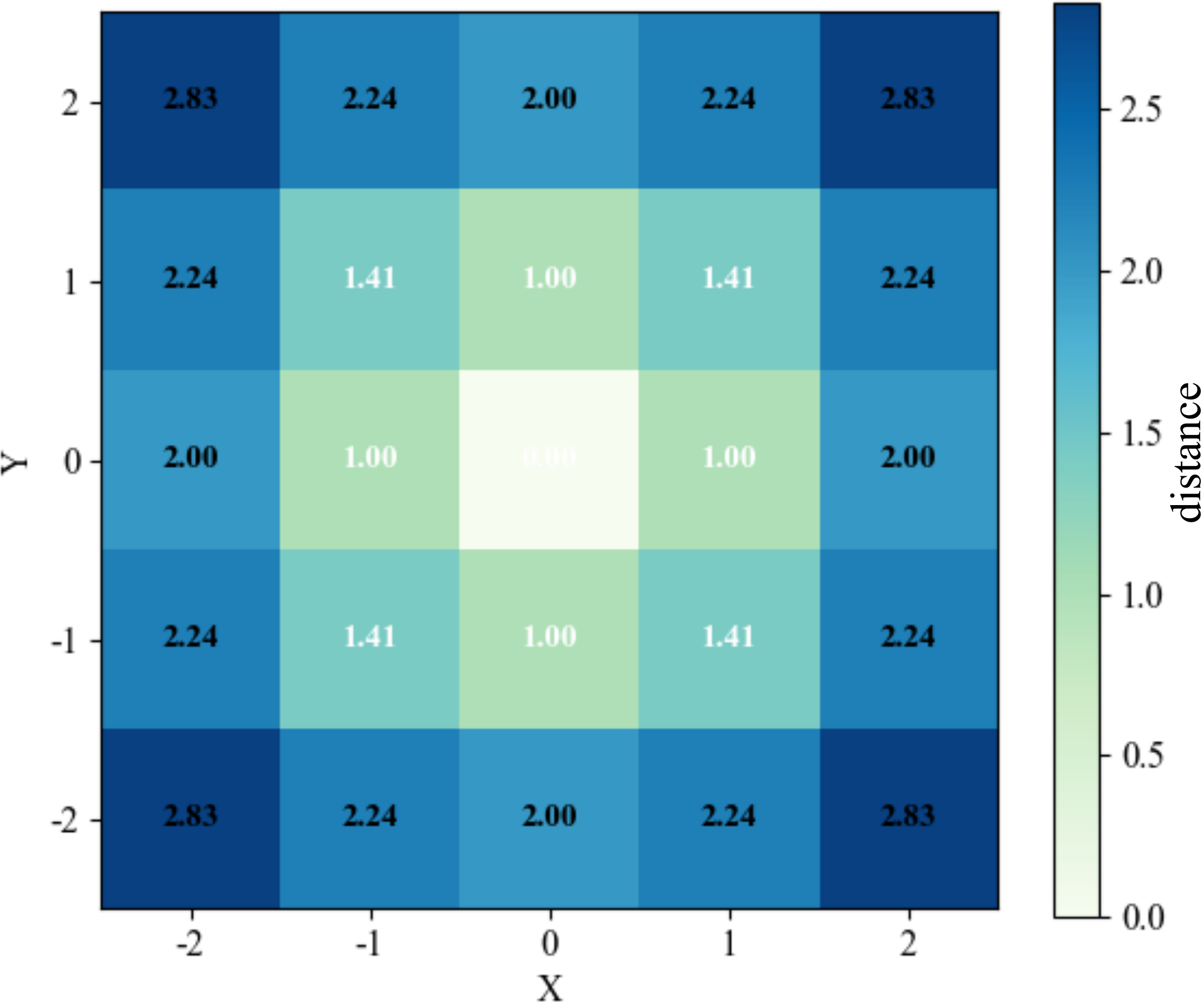}\\[2pt]{\scriptsize (a)}
    \end{minipage}\hfill
    \begin{minipage}{0.22\textwidth}\centering
    \includegraphics[width=\textwidth]{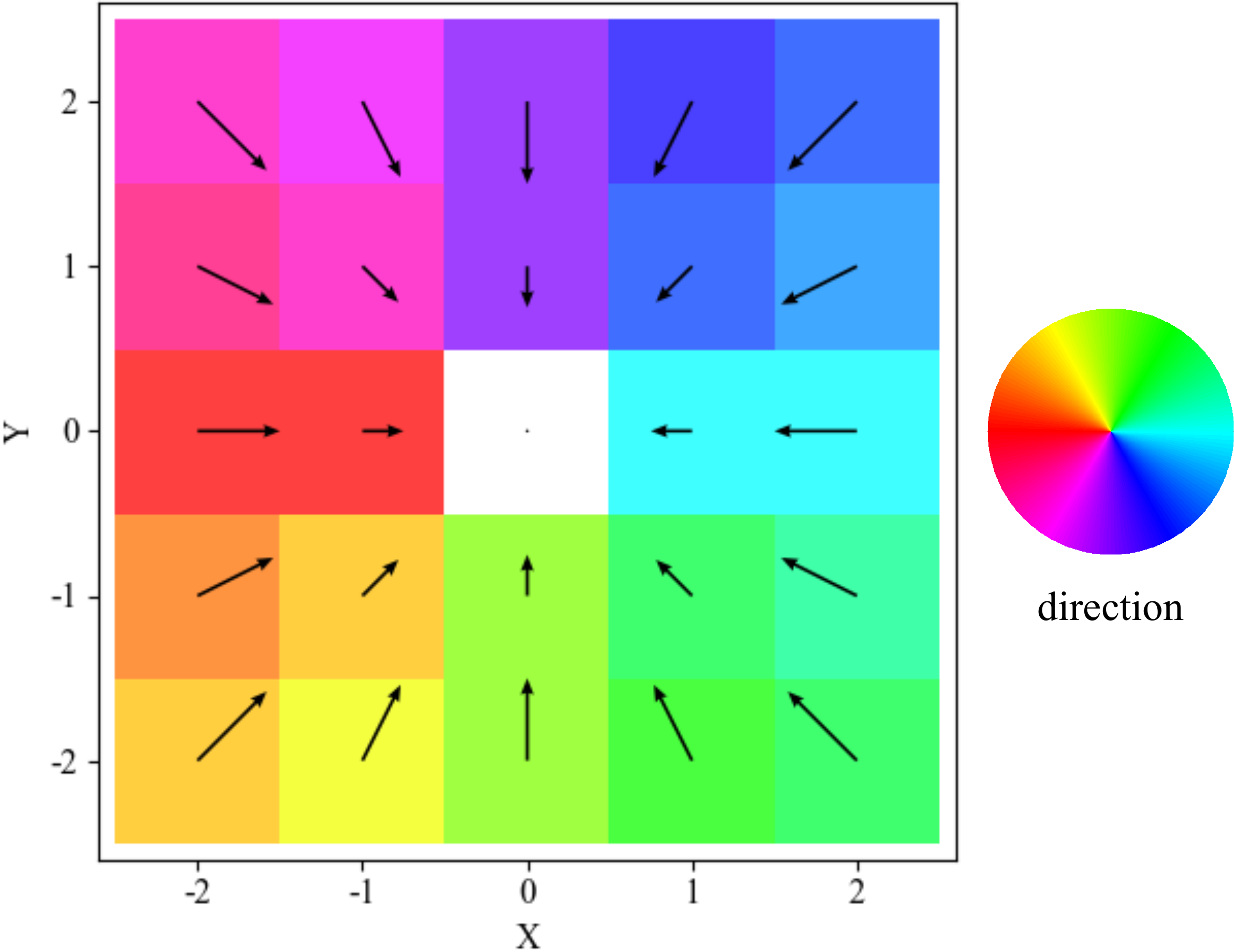}\\[2pt]{\scriptsize (b)}
    \end{minipage}

    \vspace{4pt}

    \begin{minipage}{0.22\textwidth}\centering
    \includegraphics[width=\textwidth]{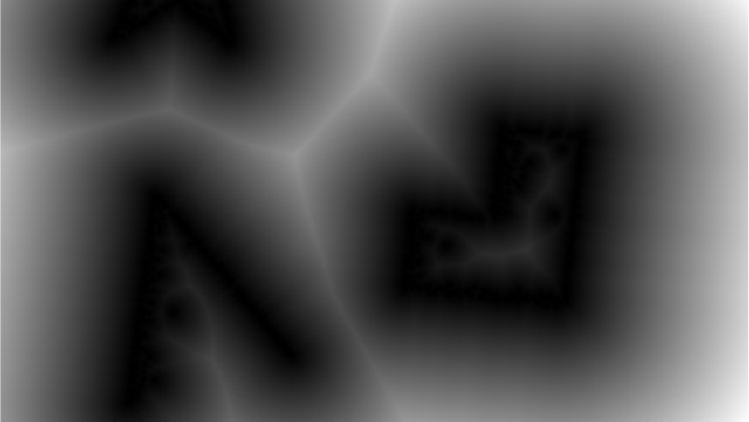}\\[2pt]{\scriptsize (c)}
    \end{minipage}\hfill
    \begin{minipage}{0.22\textwidth}\centering
    \includegraphics[width=\textwidth]{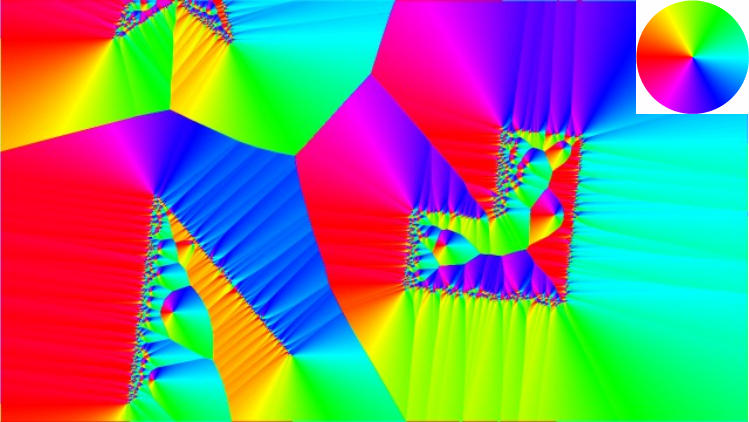}\\[2pt]{\scriptsize (d)}
    \end{minipage}
    \caption{\Add{Distance field and direction field. (a-b) Toy example around a single edge point, showing how the distance vector at each location encodes translation magnitude and direction. (c-d) The same fields computed on a real graphics scene, used to precompute the distance vectors that give the translation direction and amount for subsequent events.}}
    \label{fig:distance and direction for grid}
    \vspace{-17pt}
\end{figure}

\Add{Let $\hat{d}(x) = -\nabla D(x)/\|\nabla D(x)\|$ denote the unit gradient-descent direction of the field at pixel $x$. The translation vector implied by $x$, referred to elsewhere in this paper simply as its distance vector, is then $v(x) = D(x)\,\hat{d}(x)$, precomputed and stored as the pair $(D,\hat d)$ (Fig.~\ref{fig:distance and direction for grid}). Each new event reads $v(x)$ directly from these fields, and averaging $v(x)$ across the current batch gives the displacement that shifts the field forward.}

\begin{figure}[!t]
    \centering
    \begin{minipage}{0.113\textwidth}\centering
    \includegraphics[width=\textwidth]{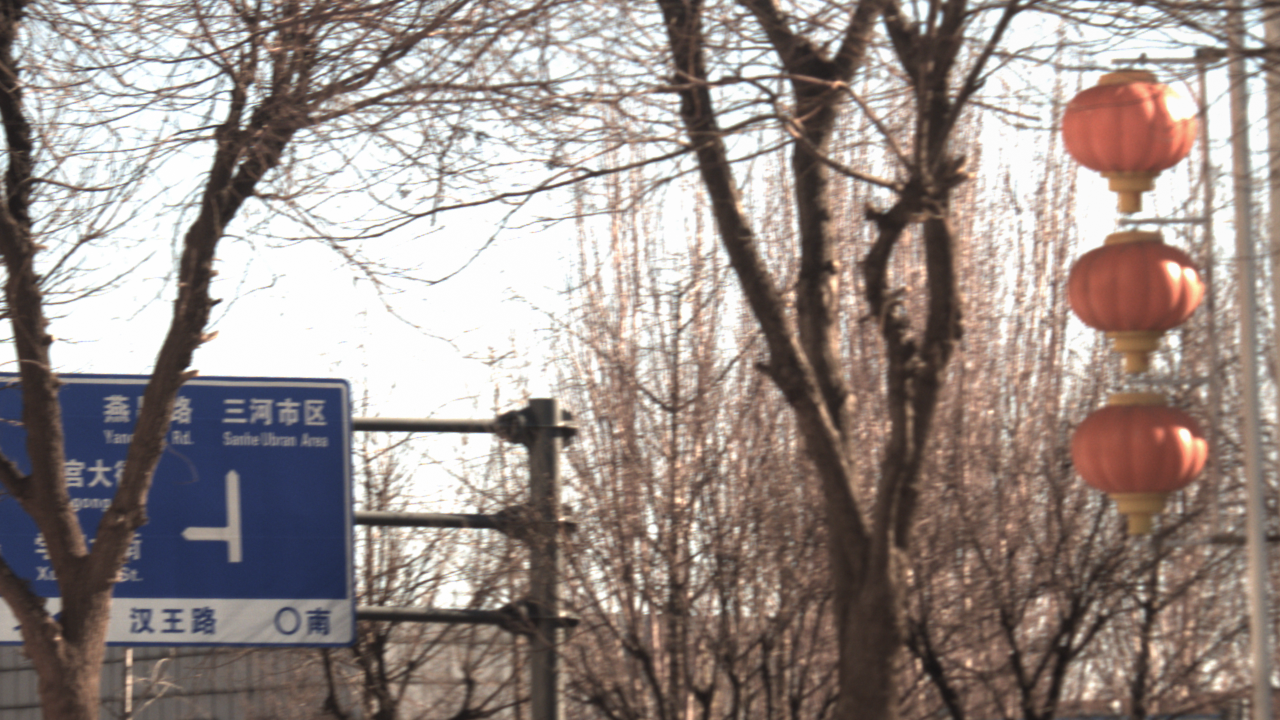}\\[2pt]{\scriptsize (a)}
    \end{minipage}\hspace{1mm}
    \begin{minipage}{0.113\textwidth}\centering
    \includegraphics[width=\textwidth]{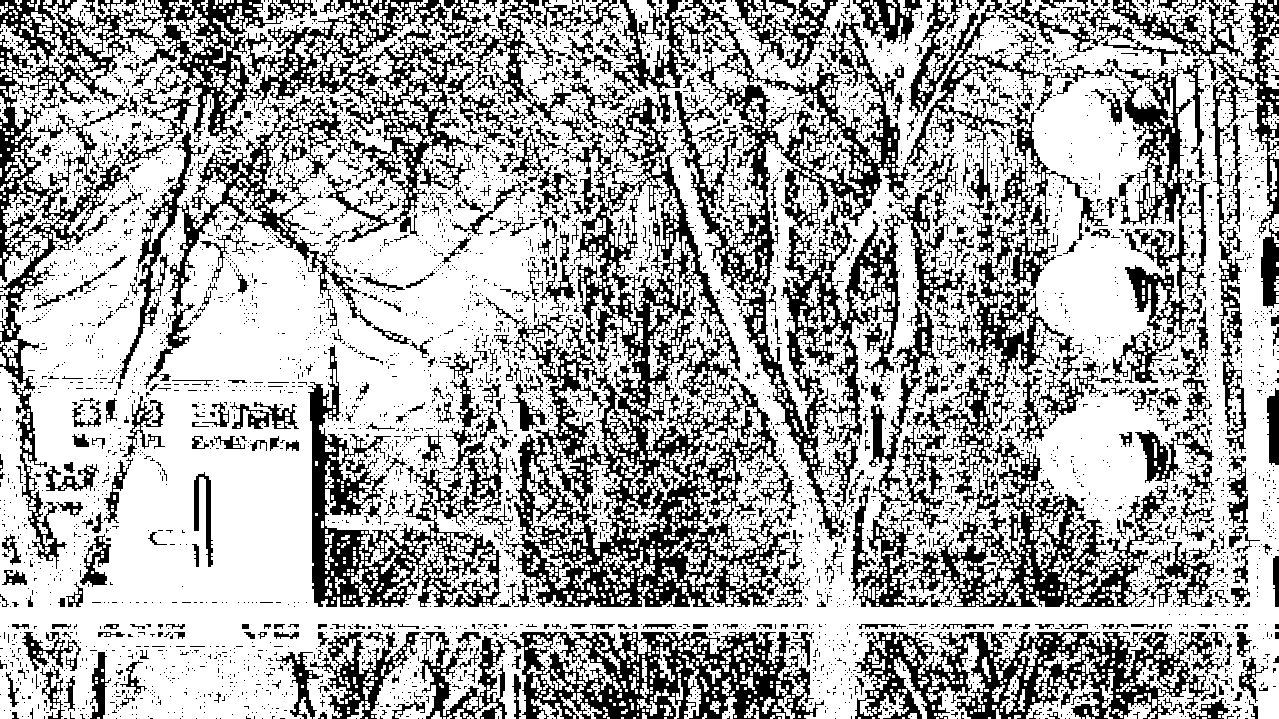}\\[2pt]{\scriptsize (b)}
    \end{minipage}\hspace{1mm}
    \begin{minipage}{0.113\textwidth}\centering
    \includegraphics[width=\textwidth]{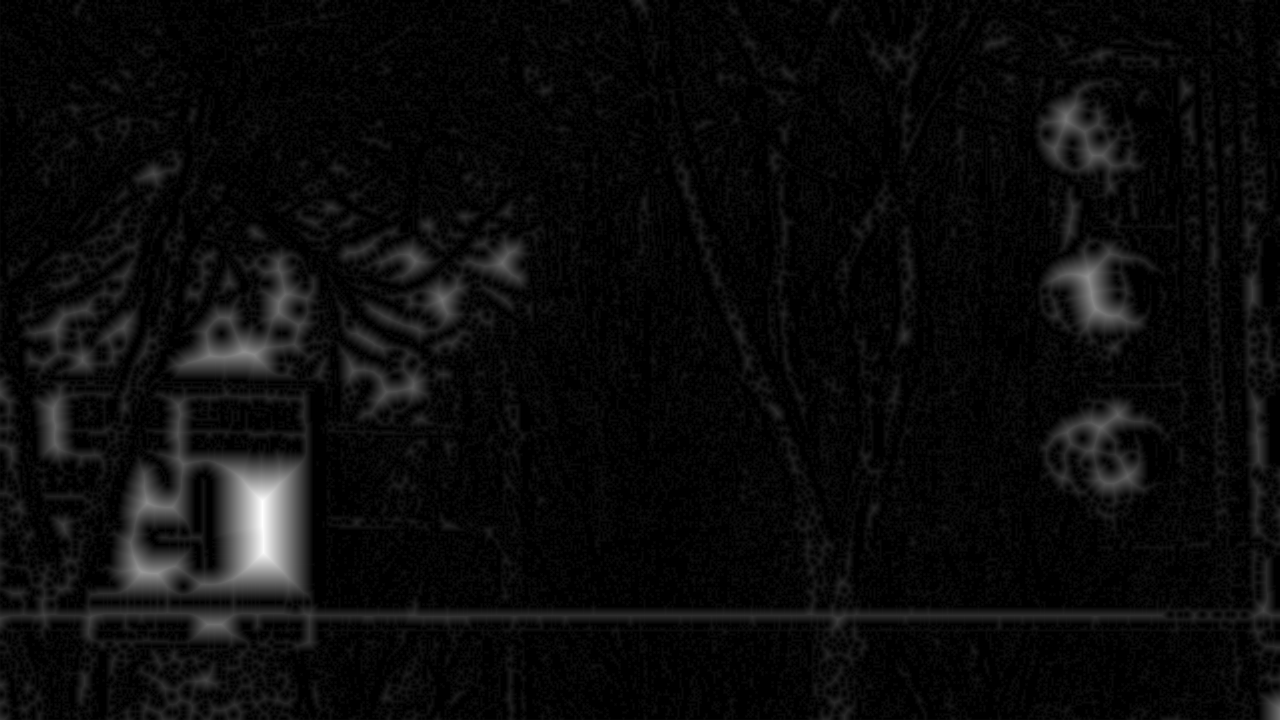}\\[2pt]{\scriptsize (c)}
    \end{minipage}\hspace{1mm}
    \begin{minipage}{0.113\textwidth}\centering
    \includegraphics[width=\textwidth]{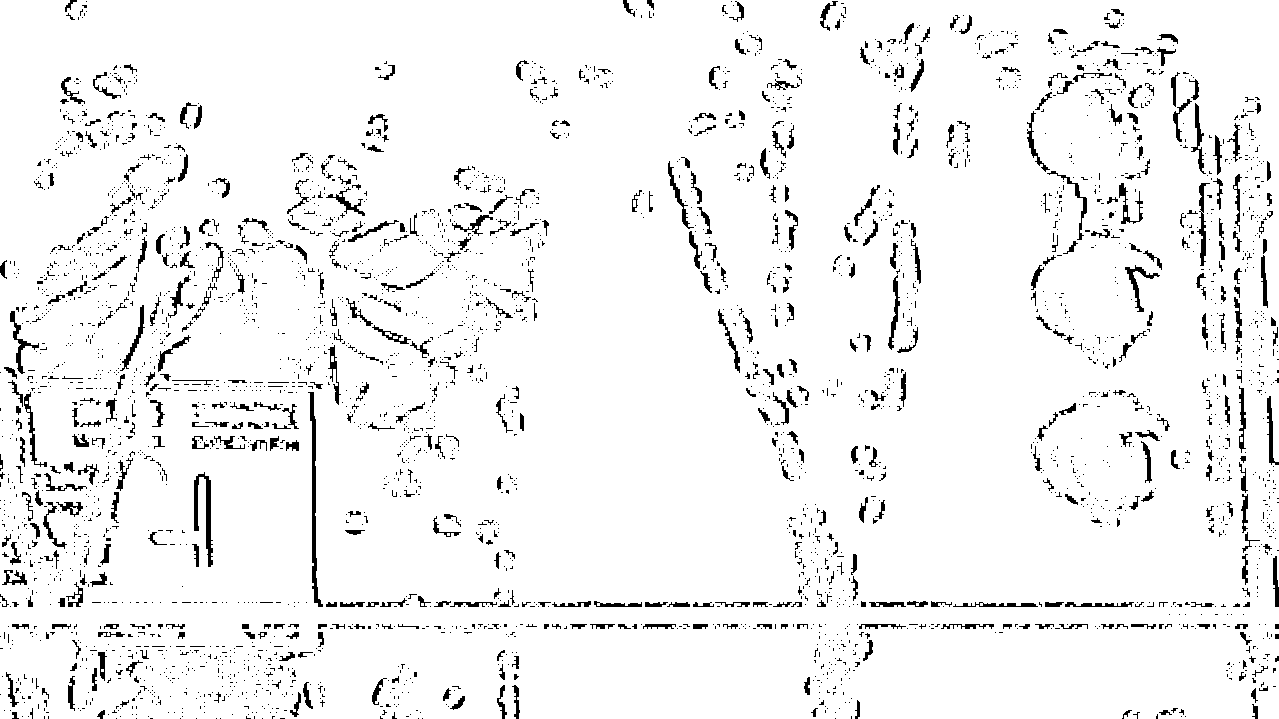}\\[2pt]{\scriptsize (d)}
    \end{minipage}
    \caption{\Add{Valid edge region selection. (a) reference image. (b) edge template. (c) distance map from the edge template. (d) valid region after dilating sparse-edge regions and intersecting with the edge active region.}}
    \label{fig:valid_edge}
    \vspace{-17pt}
\end{figure}

\Add{A single event, however, cannot fully determine $t$: picture a straight edge sliding behind a small aperture, where motion toward or away from the edge's line is visible, but sliding along its direction leaves the view unchanged. This is the classical aperture problem, and it applies to $v(x)$ identically, since reading $v(x)$ from one event constrains only the component of $t \in \mathbb{R}^2$ along that event's local edge normal. Orientation diversity within a batch resolves the ambiguity: edges of differing orientation constrain different components of $t$, so their normal components reinforce each other under averaging while no single direction dominates, a good approximation precisely when a batch's orientations are reasonably spread, and unreliable parallel to a template's dominant orientation otherwise, for instance one long straight boundary, a failure mode we return to in Sec.~\ref{sec:conclusion}. A separate problem arises in edge-dense regions, where field values are small everywhere and a raw average is biased toward zero regardless of orientation diversity. We correct this by dilating the sparse-edge regions where the field exceeds 5~pixels and intersecting the result with the edge active region, restricting the average to this reduced valid region (Fig.~\ref{fig:valid_edge}). Both thresholds are empirical, chosen by inspecting when the dense-region bias became visually apparent, and are not delicate to tune since they only need to separate this small-value regime from the larger, informative values elsewhere.}

\Add{Events generated far from any edge are noise that would otherwise pull the average off the true trajectory, so we discard events whose distance from the edge exceeds this same 5-pixel scale, since genuine correspondences rarely deviate this far under our sensors' noise and quantization, while spurious events, not concentrated at any particular distance, are thinned out by the cutoff.}

\subsection{Extracting Trajectory}

\Add{We now recover the exposure-interval trajectory from the per-event translation vectors $v(x)$ of Sec.~\ref{subsec:field_generate}. As noted in~\cite{gao2024n}, a single readout only recovers the component of motion perpendicular to the local edge, so we average $v(x)$ over a batch of $M$ events, $\bar v = \frac{1}{M}\sum_{i=1}^{M} v(x_i)$, to approximate the full 2-DoF displacement via the orientation-diversity argument of Sec.~\ref{subsec:field_generate}.}

\Add{Choosing $M$ matters: large batches blur the translation estimate, while small batches let event noise dominate and the trajectory oscillate. We set $M$ to 2.5\% of the event count used to build the edge template (Sec.~\ref{subsec:pattern_generation}), empirically chosen since we lack a closed-form model of how noise scales with batch size, by observing where the trajectory stopped oscillating without yet showing visible blur. We have not formally swept its sensitivity, but the two failure modes are visually distinct enough that the setting should not be delicately poised between them.}

\Add{Trailing events left behind by the moving edge likewise blur $\bar v$, which we suppress with a Spatio-Temporal-Contrast filter for event-stream denoising before averaging.}

\Add{The full trajectory $s(t)$ is reconstructed by processing events sequentially within the exposure interval, concatenating each batch's incremental displacement $\bar v$. This process is non-iterative, so each incremental estimate carries some error, but the error does not compound, since every $\bar v$ is read from the field anchored to the fixed initial template rather than accumulated relative to the previous, already noisy estimate.}

\noindent\textbf{Discussion.}
\Add{The efficiency of ODF motion estimation follows from where its computation is spent, not a constant-factor engineering optimization. Building the distance and direction fields over an $H \times W$ template costs $O(HW)$, paid once per template rather than once per event: every subsequent event then costs only a field lookup and an accumulation into a running average, independent of scene complexity or motion magnitude. An iterative solver instead pays a cost that depends on how many refinement steps convergence takes, and a multi-hypothesis tracker pays a cost proportional to the states it compares per event. Memory follows the same pattern, two fixed $O(HW)$ fields per template rather than a hypothesis set or optimizer state that grows with tracking duration. ODF motion estimation's per-event cost is thus decoupled from motion magnitude, scene texture, and event count, the very factors that determine the cost of optimization- or search-based alternatives, a property Sec.~\ref{sec:experiments} validates through update rate and latency.}

\section{Applications}
\label{sec:applications}

\subsection{Image Deblurring with Estimated Kernel}
\label{sec:deblurring}

\subsubsection{Kernel Generator}

The motion blur point spread function (PSF) follows from the trajectory $s(t)$ over exposure time $T$~\cite{boracchi2012modeling}:
\begin{equation}
    PSF(\cdot) = \int_0^T \delta_{s(t)}(\cdot)dt,
\end{equation}
where $\delta_{s(t)}$ is the Dirac delta function at $s(t)$.

To build the kernel, each position along the trajectory is assigned a value equal to the fraction of exposure time spent there, with sub-pixel linear interpolation at fractional positions. Unlike typical 8-bit kernels, we keep it in continuous, unquantized form throughout non-blind deblurring.

\Add{Because the clean-image ground truth in most deblurring datasets corresponds to the exposure's middle rather than the kernel's geometric center, we center the kernel on the trajectory position at the temporal midpoint of exposure, which ODF motion estimation gives directly since it recovers the exact pixel position at any instant during exposure.}

\subsubsection{Deep Unfolding Network for Deblurring with Noisy Blur Kernels}
\label{sec:edi}

Event noise, sensor bandwidth limits, and kernel quantization mean the estimated kernel inevitably deviates from the ground truth. Prior noise models for non-blind deblurring~\cite{vasu2018non,dong2021dwdn} assume error distributions inconsistent with ours and do not transfer directly.

\Add{Inspired by the half-quadratic splitting scheme in USRNet~\cite{zhang2020deep} and DPIR~\cite{zhang2022plugandplay}, we build an iterative deep-unfolding network that alternates a data module with a learned prior module. Given the blur kernel $k$, the blurred image $y$, and the prior module's previous output $z_{i-1}$, the data module solves the data-fidelity term in closed form,
\begin{equation}
    x_i = \mathcal{F}^{-1}\left(\frac{\overline{\mathcal{F}(k)}\mathcal{F}(y) + \alpha_i \mathcal{F}(z_{i-1})}{\overline{\mathcal{F}(k)}\mathcal{F}(k) + \alpha_i}\right),
\end{equation}
where $\mathcal{F}$ is the Fourier transform, the overline denotes complex conjugation, and $\alpha_i$ is a per-iteration regularization weight. The prior module then denoises $x_i$ into $z_i$, conditioned in USRNet and DPIR on a known noise level unavailable here since the true kernel-estimation noise is not directly observable. We instead condition on the original blurred image, which carries less noise than the network's own intermediate estimate, together with an event mask, a binary map marking pixels where the event count during exposure exceeds a threshold. Since these are the only pixels where the blurred and sharp image can differ, the mask focuses the network's correction on genuinely blurred regions.}

\subsubsection{Training on Synthetic Data}

Replaying synthetic trajectories through an event simulator would give blur kernels with realistic estimation noise, but current simulators emit frame-locked event timing that biases the estimated kernel. We instead generate motion-noise-infused blur kernels directly from the constructed trajectories without simulating events, convolving a low-quality image with the ground-truth kernel and noise to form the network input, supervised by the original high-quality image.

\Add{To avoid cyclic-boundary artifacts, we blur an oversized patch before cropping to the training resolution.} Kernel size varies from 11 to 81 pixels across batches using diverse motion trajectories~\cite{boracchi2012modeling}, with noise from simplified BSRGAN~\cite{Zhang_2021_ICCV} and Real-ESRGAN~\cite{Wang_2021_ICCV} pipelines and simulated pixel saturation.

\subsection{Pupil Motion Estimation for Eye Tracking}
\label{sec:eye_tracking}

\Add{Eye tracking estimates the visual axis, most commonly by locating the pupil center together with one or more corneal reflection glints under the PCCR framework~\cite{guestrin2006general}. The pupil and glints change shape as the eye rotates, so ODF motion estimation (Sec.~\ref{sec:motion_estimation}) does not apply directly, but the translation vectors it introduces still generalize to event-based spatial filtering of these features.}

\subsubsection{Pupil and Glint Event Signatures}
\Add{An initial frame localizes the pupil ellipse and the corneal glints, and the subsequent event stream tracks both asynchronously. Because the pupil is darker than the surrounding iris, its motion generates negative events on the leading edge and positive events on the trailing edge, while the smaller, brighter glints generate the opposite pattern. As in Sec.~\ref{sec:motion_estimation}, only events near the previous feature boundary are processed.}

\subsubsection{Distance-Vector-Based Pupil Event Filtering}
\Add{Pupil events are corrupted by trailing events left behind by the moving edge and by events from nearby glints. A fixed-polarity trail filter fails whenever the same pixel crosses two edges of the same sign in a row (Fig.~\ref{fig:trail}(a,b)), which happens routinely as the pupil moves. We instead estimate the pupil motion direction by summing, over nearby events, the vector from each event to the previous pupil center, with negative-event vectors flipped in sign so both polarities reinforce the same direction. An event is discarded as a trail event when its distance vector against the previous pupil ellipse (Sec.~\ref{subsec:field_generate}) opposes this direction (Fig.~\ref{fig:trail}(c)). Events inside the glint template (Sec.~\ref{subsec:glint}) are filtered separately.}

\subsubsection{Pupil Tracking}
\begin{figure}[!t]
    \centering
    \begin{minipage}{0.093\textwidth}\centering
    \includegraphics[width=\textwidth]{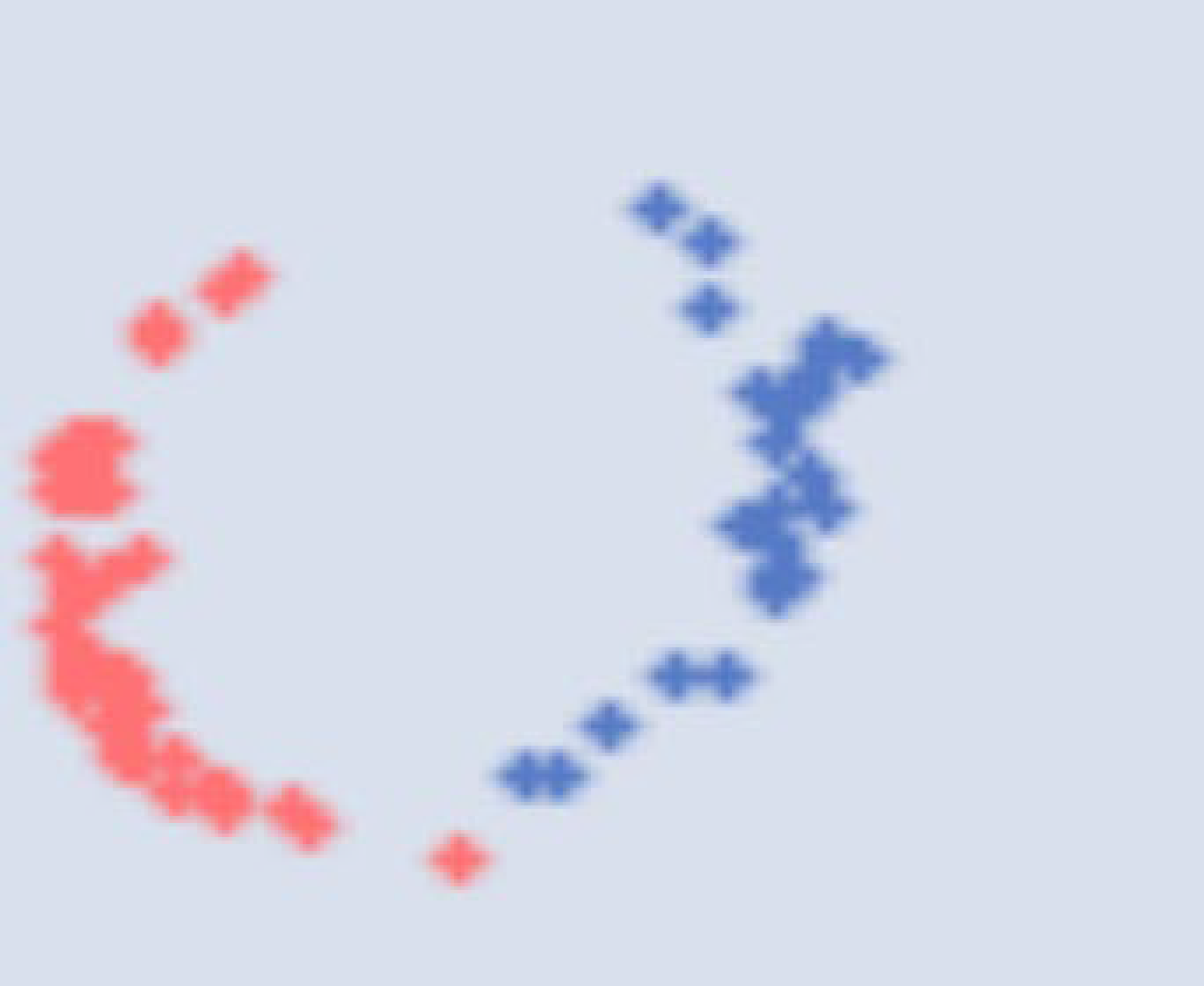}\\[2pt]{\scriptsize (a)}
    \end{minipage}\hfill
    \begin{minipage}{0.093\textwidth}\centering
    \includegraphics[width=\textwidth]{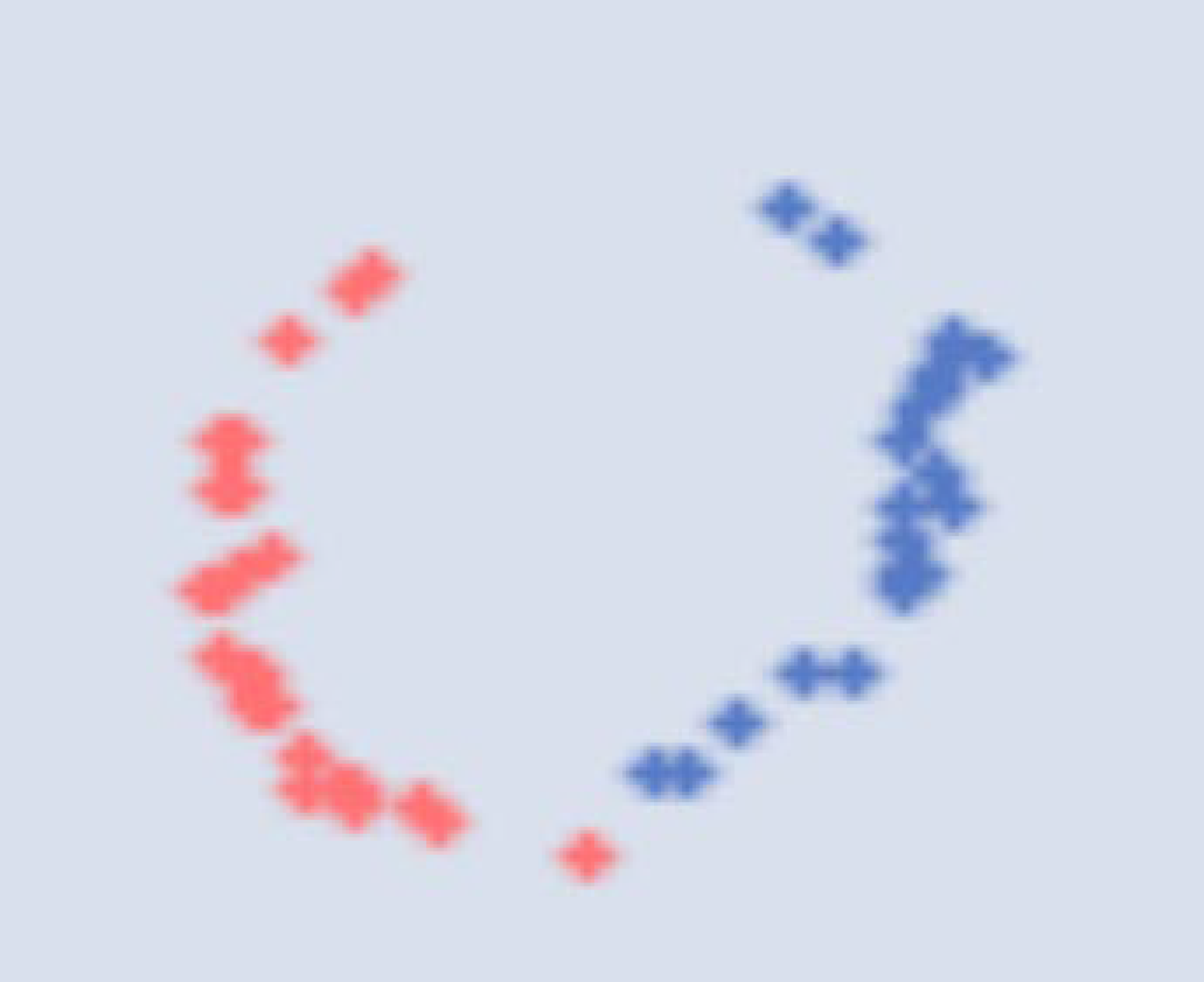}\\[2pt]{\scriptsize (b)}
    \end{minipage}\hfill
    \begin{minipage}{0.093\textwidth}\centering
    \includegraphics[width=\textwidth]{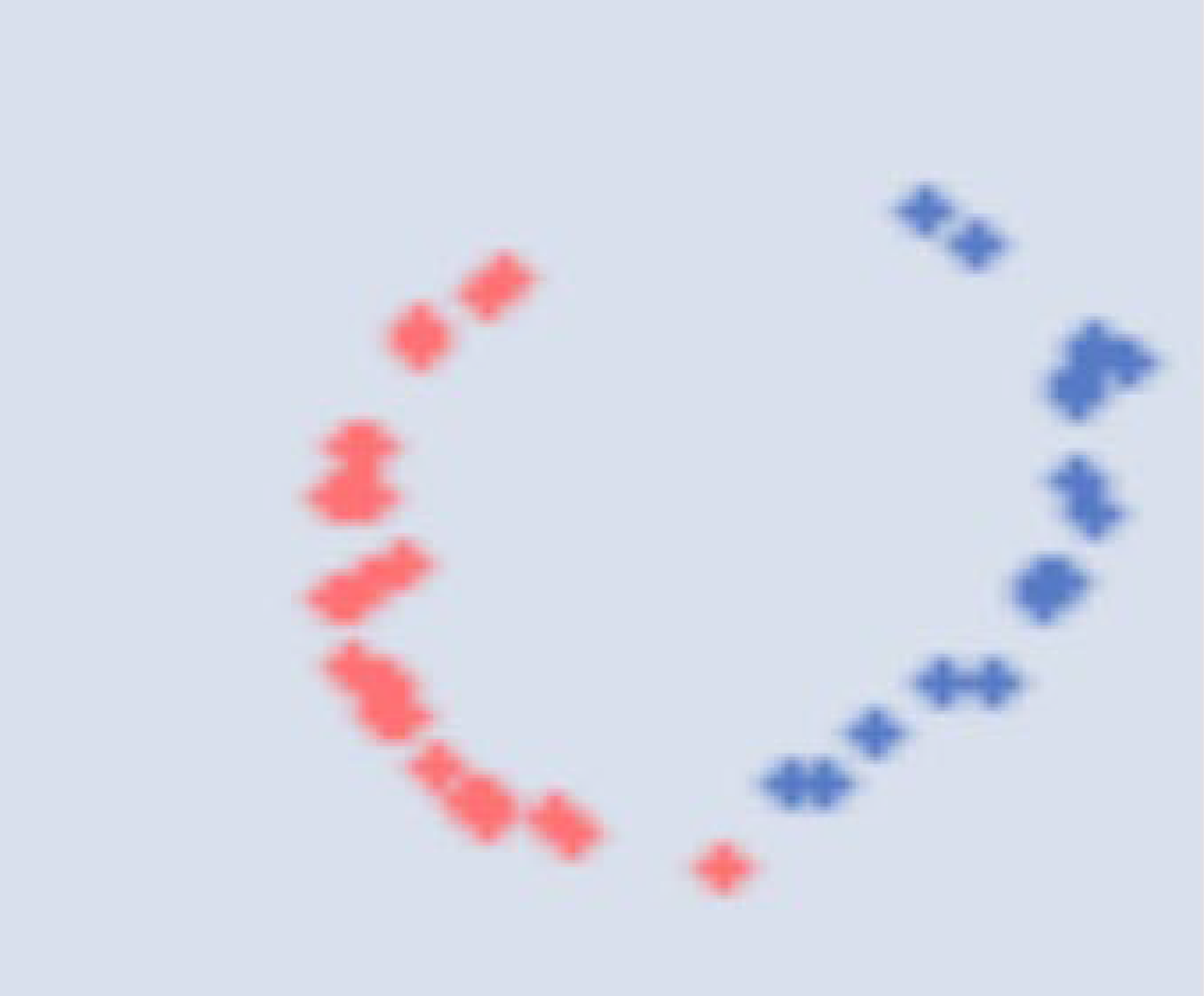}\\[2pt]{\scriptsize (c)}
    \end{minipage}\hfill
    \begin{minipage}{0.093\textwidth}\centering
    \includegraphics[width=\textwidth]{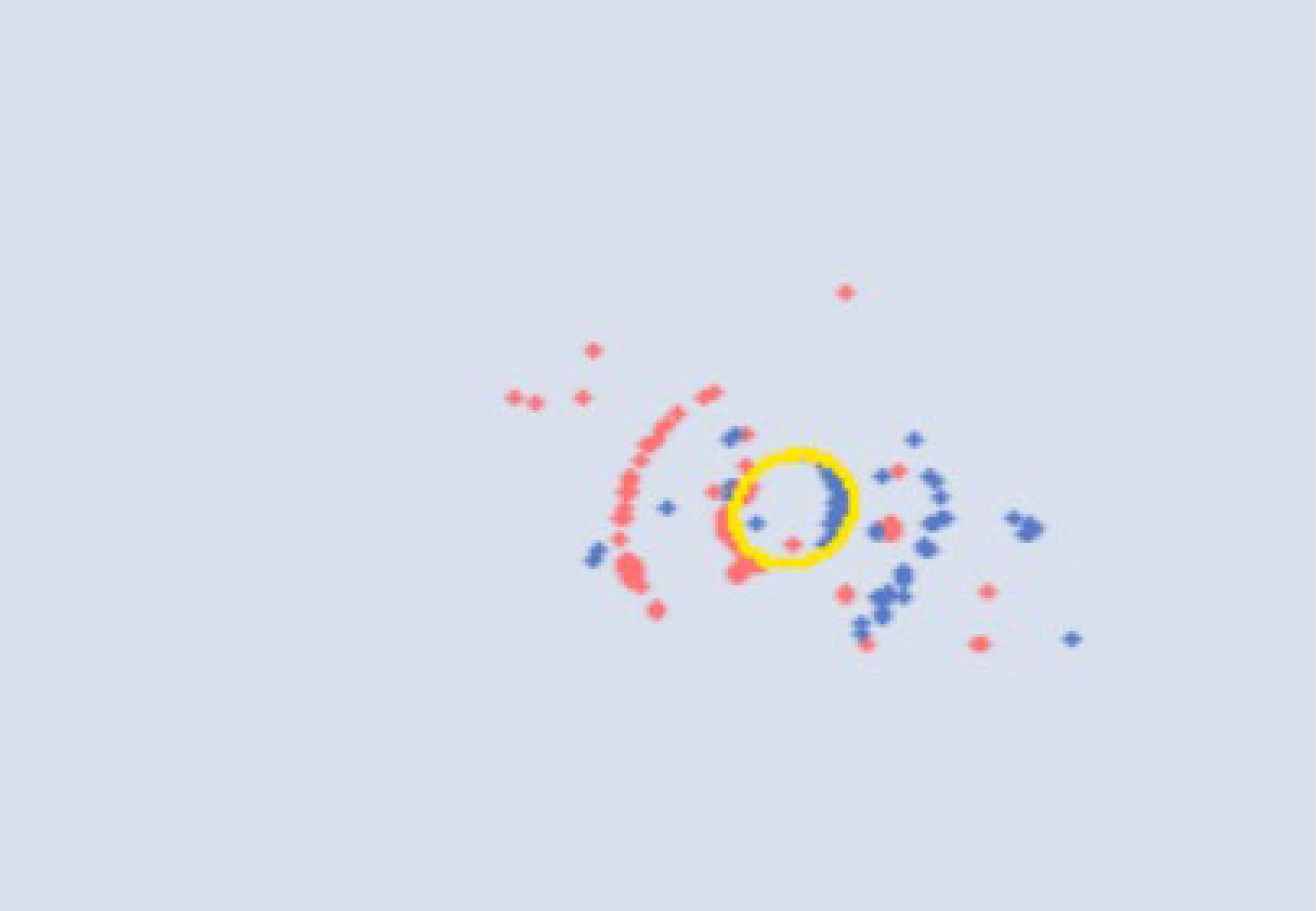}\\[2pt]{\scriptsize (d)}
    \end{minipage}\hfill
    \begin{minipage}{0.093\textwidth}\centering
    \includegraphics[width=\textwidth]{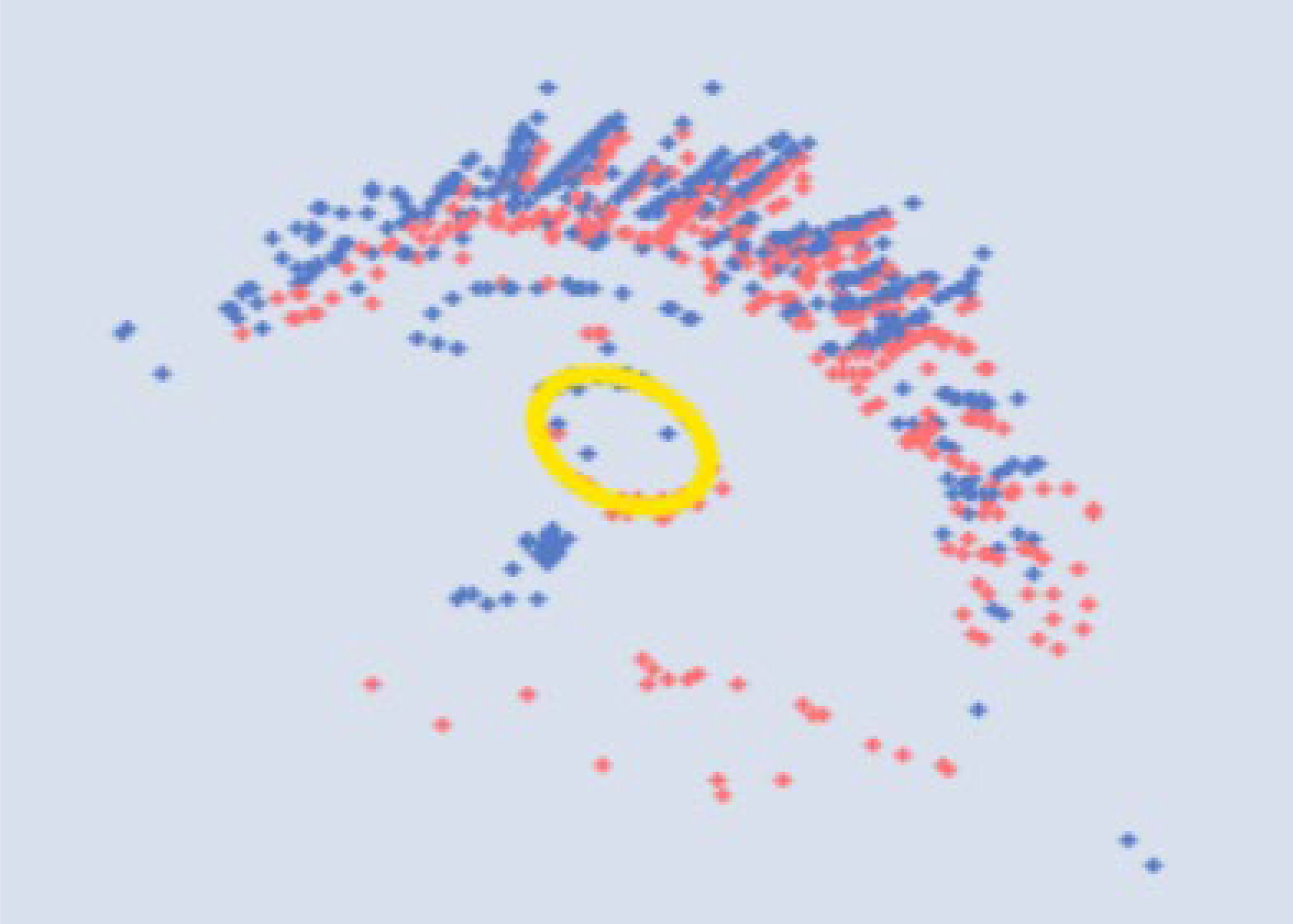}\\[2pt]{\scriptsize (e)}
    \end{minipage}
    \caption{\Add{Pupil event filtering and device comparison. (a) Raw events during pupil tracking. (b) Valid events after a fixed-polarity trail filter, which still leaks trailing events. (c) Valid events after the proposed distance-vector direction filter. (d) Pupil events captured by a non-near-eye device, where the pupil occupies few pixels. (e) Pupil events captured by a near-eye device, where the pupil occupies many more pixels.}}
    \label{fig:trail}
    \vspace{-17pt}
\end{figure}

\Add{Valid events are treated as new boundary points and refit to an ellipse with direct least-squares fitting, updated every fixed number of events, tuned to the initial ellipse size and to whether the device is near-eye (Fig.~\ref{fig:trail}(d,e)) to balance fit stability against tracking drift. Points sampled from the previous ellipse boundary are added to each new fit to further stabilize it against event noise, removing short-axis oscillation and tracking failures.}

\subsubsection{Glint Localization and Blink Detection}
\label{subsec:glint}
\Add{Because glint polarity is reversed relative to the pupil, a directional template split at the axis perpendicular to the pupil motion direction ($0$ on the axis, $+1$/$-1$ on each side) is accumulated event by event to score candidate glint regions. The accumulator is thresholded and decayed rather than reset, since glints move less than the pupil and can disappear intermittently. Blinks are detected from an abrupt, sustained drop in the pupil ellipse's minor axis, which suspends tracking.}

\section{Experiments}
\label{sec:experiments}

\subsection{Event-Based Motion Estimation}
\Add{We evaluate motion estimation accuracy and update rate against contrast maximization (CM)~\cite{gallego2018unifying} and HASTE~\cite{alzugaray:BMVC20}, both pure-event methods, on four Event-Camera Dataset~\cite{mueggler2017event} sequences where a motorized slider translates the camera along one axis (\textit{slider\_close}, \textit{slider\_far}, and their high-dynamic-range counterparts). Motion is horizontal, so every method is constrained to 1 degree of freedom to avoid the aperture problem of Sec.~\ref{sec:motion_estimation}. ODF motion estimation and HASTE build their edge template with the adaptive event-count selection of Sec.~\ref{subsec:pattern_generation}. For CM, we sweep its energy function, event-count window, and blur parameter and report the best configuration. Ground truth comes from the slider's timestamped physical displacement.}

\begin{table}[!t]
\caption{\Add{Motion estimation accuracy and update rate on Event-Camera Dataset~\cite{mueggler2017event} (bold: best per column).}}
\vspace{-11pt}
\centering
\small
\resizebox{0.48\textwidth}{!}{
\setlength{\tabcolsep}{8pt}
\renewcommand\arraystretch{1.0}
\begin{tabular}{l l c c}
\bottomrule[0.15em]
\rowcolor{tableHeadGray}
\textbf{Sequence} & \textbf{Method} & \textbf{Rate $\uparrow$ (Hz)} & \textbf{Error $\downarrow$ (px)} \\ \hline
\multirow{3}{*}{slider\_close} & HASTE~\cite{alzugaray:BMVC20} & 467.2 & \textbf{0.396} \\
 & CM~\cite{gallego2018unifying} & 19.5 & 0.734 \\
 & Ours & \textbf{1066.0} & 0.584 \\ \hline
\multirow{3}{*}{slider\_far} & HASTE~\cite{alzugaray:BMVC20} & 188.9 & 0.457 \\
 & CM~\cite{gallego2018unifying} & 17.2 & 0.558 \\
 & Ours & \textbf{679.4} & \textbf{0.328} \\ \hline
\multirow{3}{*}{slider\_hdr\_close} & HASTE~\cite{alzugaray:BMVC20} & 632.1 & 0.555 \\
 & CM~\cite{gallego2018unifying} & 15.5 & \textbf{0.426} \\
 & Ours & \textbf{1036.5} & 0.543 \\ \hline
\multirow{3}{*}{slider\_hdr\_far} & HASTE~\cite{alzugaray:BMVC20} & 175.9 & 0.496 \\
 & CM~\cite{gallego2018unifying} & 12.2 & 0.564 \\
 & Ours & \textbf{533.9} & \textbf{0.381} \\ \hline
\end{tabular}
}
\vspace{-15pt}
\label{table:motion_estimation}
\end{table}

\Add{Table~\ref{table:motion_estimation} shows all three methods reaching sub-pixel error, confirming that a closed-form average does not by itself sacrifice accuracy. ODF motion estimation updates at the highest rate on every sequence and attains the lowest error on \textit{slider\_far} and \textit{slider\_hdr\_far}. HASTE and CM edge out a lower error on the remaining two sequences, since each can be tuned to a sequence's noise characteristics, whereas our fixed, precomputed field is not, trading the last fraction of a pixel iterative refinement can buy for an order-of-magnitude higher update rate, favorable when latency matters more than peak accuracy.}

\subsection{Image Deblurring with Estimated Kernel}

\begin{figure}[!t]
    \centering
    \includegraphics[width=0.48\textwidth]{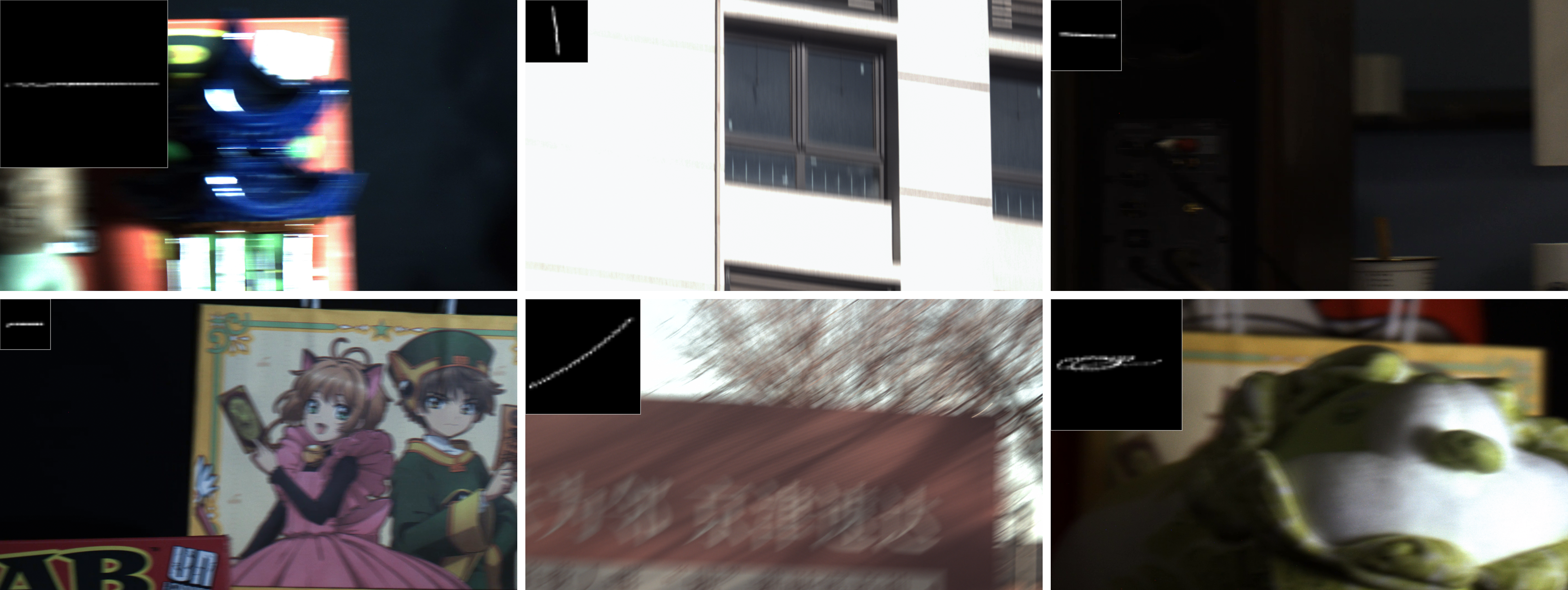}
    \caption{\Add{Blurred EventAid-B images (main) with the corresponding kernel estimated by ODF motion estimation (inset), showing the range of blur magnitude and scene content covered by the benchmark.}}
    \label{fig:kernel_and_blur}
    \vspace{-17pt}
\end{figure}

We evaluate deblurring on the global-motion sequences of EventAid-B~\cite{duan2025eventaid} (box, building, desk, jingjin, dog, global, and xiaoying), spanning the blur magnitude and scene content in Fig.~\ref{fig:kernel_and_blur} and covering degradations such as row-scanning artifacts and low-light noise, with blur from near-sharp to over 100~pixels. No method is fine-tuned on this dataset, to measure generalization to real-world data.

For kernel-accuracy evaluation, we additionally collect a real dataset of 70 blurred images with corresponding events and ground-truth blur kernels, using a synchronized, co-located event-and-frame-camera rig and a motorized stage providing the true motion trajectory from its calibrated position during exposure.

\Add{We train on 800 DIV2K~\cite{agustsson2017ntire} and 2750 Flickr2K~\cite{timofte2017ntire} images at $512\times512$ patch size, validating without test-time fine-tuning on Urban100~\cite{huang2015single} and a color image set simulated the same way. Following USRNet~\cite{zhang2020deep}, we use Adam at batch size 12, first with an $\ell_1$ loss and learning rate $1\times10^{-4}$ halved every $2\times10^5$ iterations to about $6\times10^{-6}$, then fine-tuned for $2\times10^3$ iterations with $\ell_1$, VGG perceptual, and relativistic adversarial losses weighted $[1,1,0.005]$.}

\Add{We compare against blind deblurring baselines NAFNet~\cite{chen2022simple} and Restormer~\cite{Zamir2021Restormer}, which take only the blurred image, and event-assisted baselines EDI~\cite{edi_pan} and EFNet~\cite{sun2022event} (pretrained on GoPro and, separately, fine-tuned on REBlur). To isolate ODF motion estimation's contribution from the restoration network's, we also pair our estimated kernel with two existing non-blind networks, DWDN~\cite{dong2021dwdn} and RAM~\cite{terris2025ram}, in place of our own, and separately pair it with the plain USRNet-tiny~\cite{zhang2020deep} backbone at matched parameter count to isolate our architectural modifications from the kernel and the capacity budget. All pretrained models are used without further fine-tuning on either test set, reflecting generalization rather than dataset-specific adaptation.}

\begin{table}[!t]
\caption{{Comparison of single image motion deblurring methods on EventAid-B-Global Motion. \Add{(bold: best per column)}}}
\vspace{-11pt}
\centering
\small
\resizebox{0.48\textwidth}{!}{
\setlength{\tabcolsep}{10pt}
\renewcommand\arraystretch{1.0}
\begin{tabular}{ l c  c  c  c }
\bottomrule[0.15em]
\rowcolor{tableHeadGray}
\textbf{Method} &\textbf{Events} & \textbf{PSNR} $\uparrow$ & \textbf{SSIM} $\uparrow$ & \textbf{\#Param} \\ \hline
EDI~\cite{edi_pan}                  &\cmark & 26.45 & 0.800 & - \\
Restormer~\cite{Zamir2021Restormer}          & \xmark         & 26.25  & 0.842 & 26.1M \\
NAFNet-64~\cite{chen2022simple}                 & \xmark         & 25.52  & 0.816 &  67.9M \\
EFNet-GoPro~\cite{sun2022event}                    & \cmark         & 25.65 & 0.806  & 8.5M \\
EFNet-REBlur~\cite{sun2022event}                    & \cmark         & 25.56 &  \Add{\textbf{0.874}} & 8.5M \\
ODF kernel + DWDN~\cite{dong2021dwdn} & \cmark         & 26.66  & 0.856  & 7.0M \\
ODF kernel + USRNet-tiny~\cite{zhang2020deep} & \cmark         & 26.35 &  0.843 & 0.6M \\
ODF kernel + RAM~\cite{terris2025ram} & \cmark         &  26.31 &  0.839 & 35.6M \\
\Add{\textbf{Ours}} (ODF kernel + event mask)   & \cmark         & \Add{\textbf{27.23}}  & 0.860  & 0.6M \\
\hline

\end{tabular}
}
\vspace{-15pt}
\label{table:deblur}
\end{table}

\begin{figure*}[!t]
\centering
\begin{minipage}{0.116\textwidth}\centering
\includegraphics[width=\textwidth]{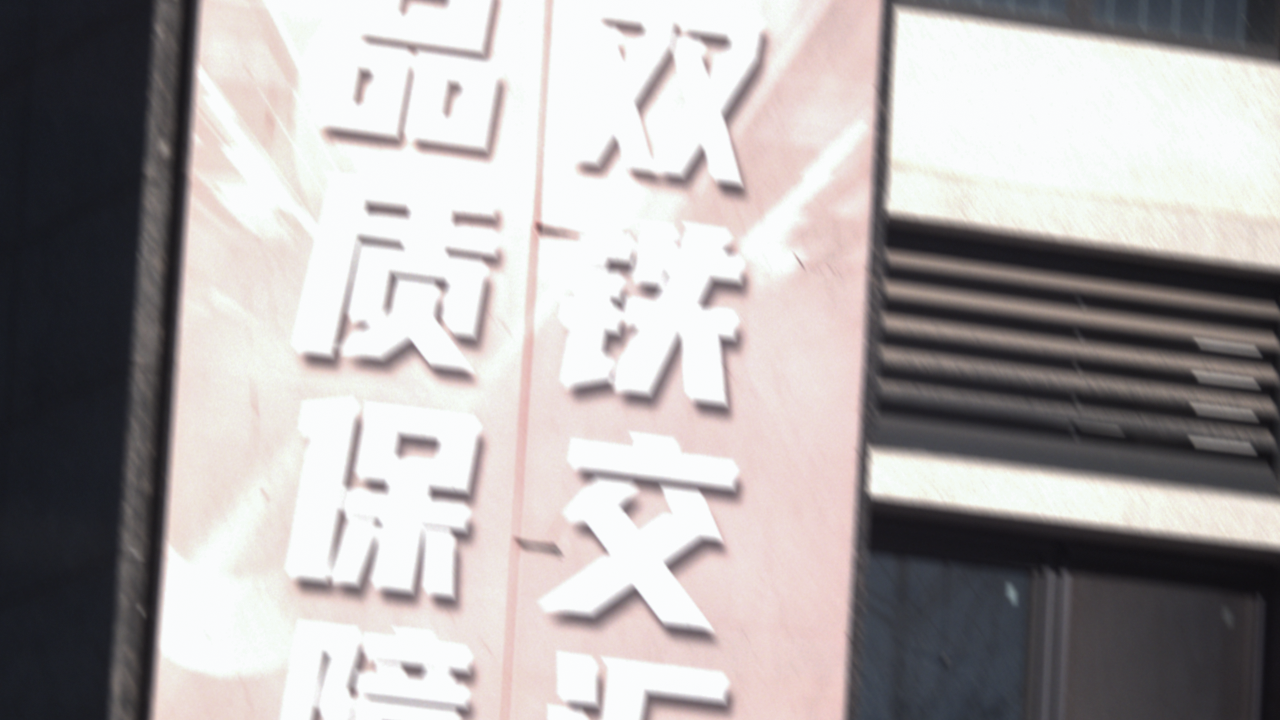}\\[2pt]{\scriptsize Blurred}
\end{minipage}\hfill
\begin{minipage}{0.116\textwidth}\centering
\includegraphics[width=\textwidth]{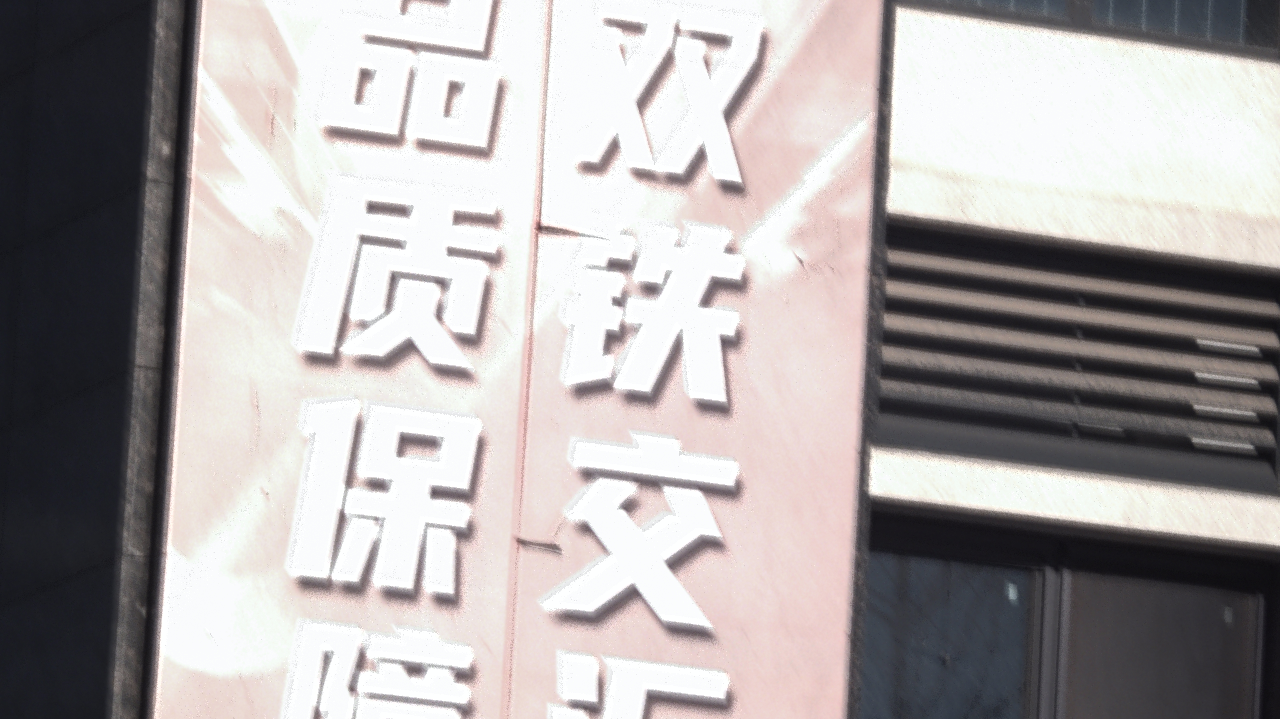}\\[2pt]{\scriptsize EDI}
\end{minipage}\hfill
\begin{minipage}{0.116\textwidth}\centering
\includegraphics[width=\textwidth]{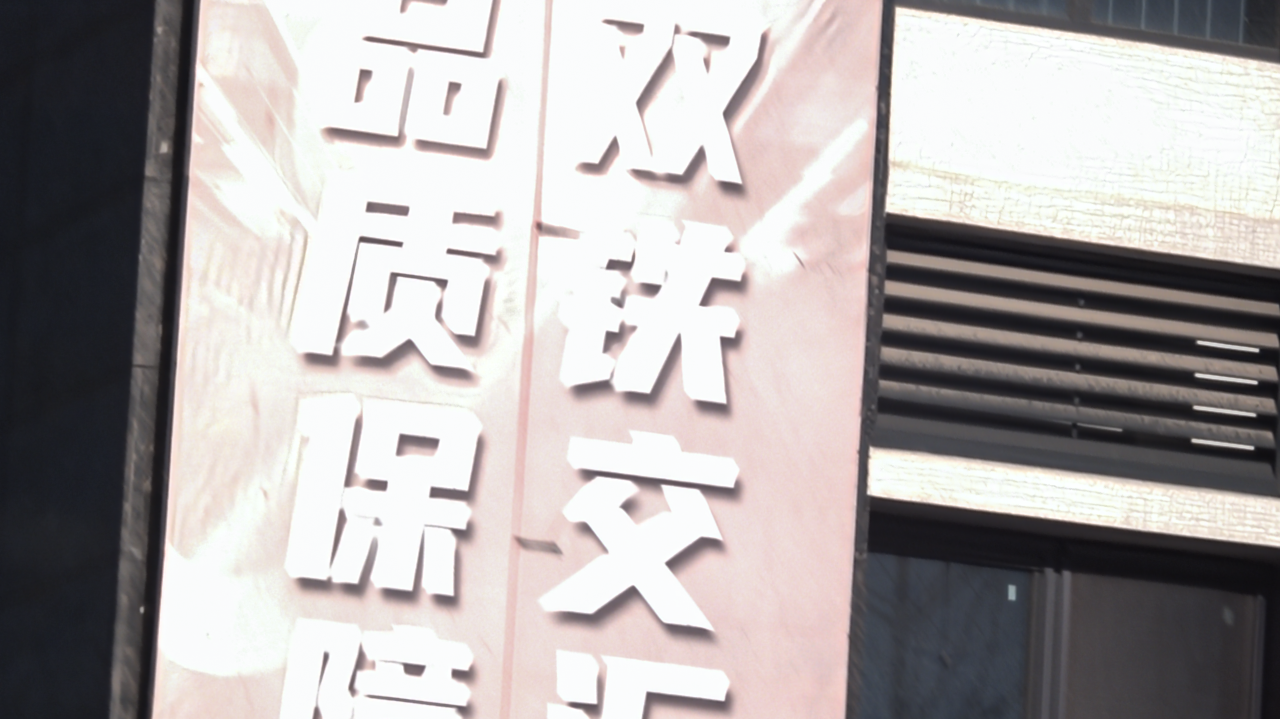}\\[2pt]{\scriptsize NAFNet}
\end{minipage}\hfill
\begin{minipage}{0.116\textwidth}\centering
\includegraphics[width=\textwidth]{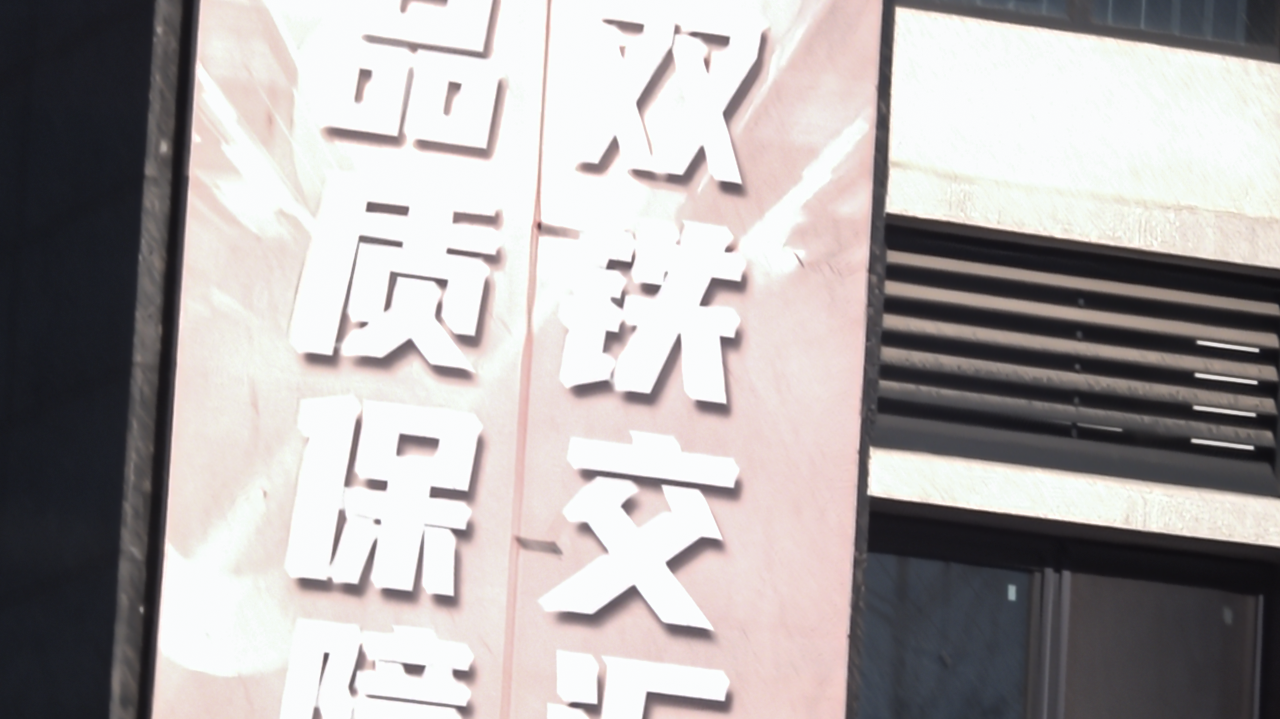}\\[2pt]{\scriptsize Restormer}
\end{minipage}\hfill
\begin{minipage}{0.116\textwidth}\centering
\includegraphics[width=\textwidth]{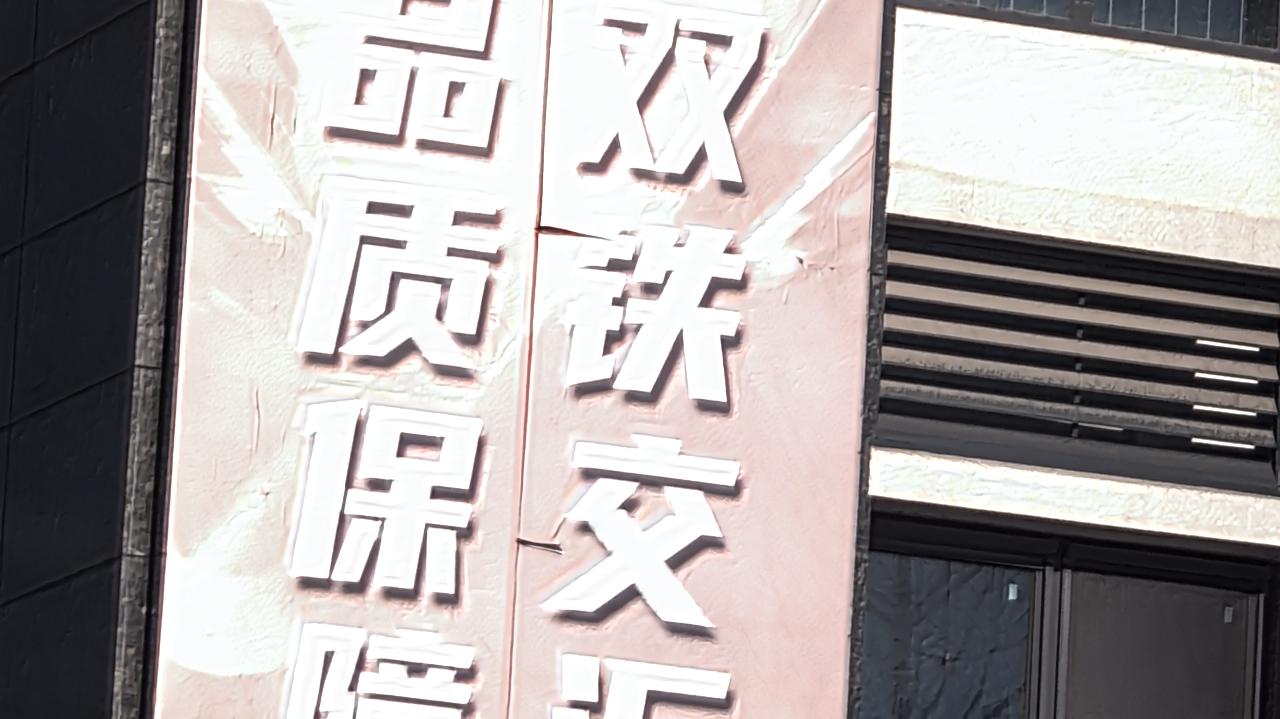}\\[2pt]{\scriptsize EFNet-GoPro}
\end{minipage}\hfill
\begin{minipage}{0.116\textwidth}\centering
\includegraphics[width=\textwidth]{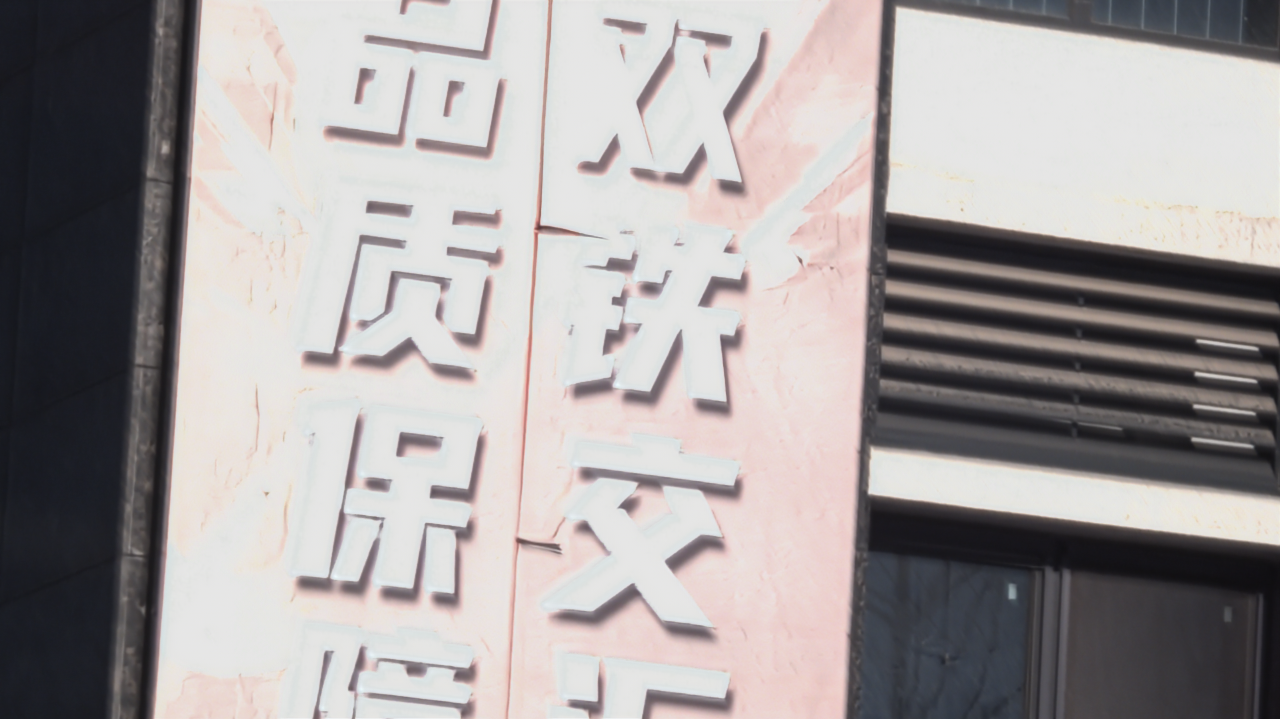}\\[2pt]{\scriptsize EFNet-REBlur}
\end{minipage}\hfill
\begin{minipage}{0.116\textwidth}\centering
\includegraphics[width=\textwidth]{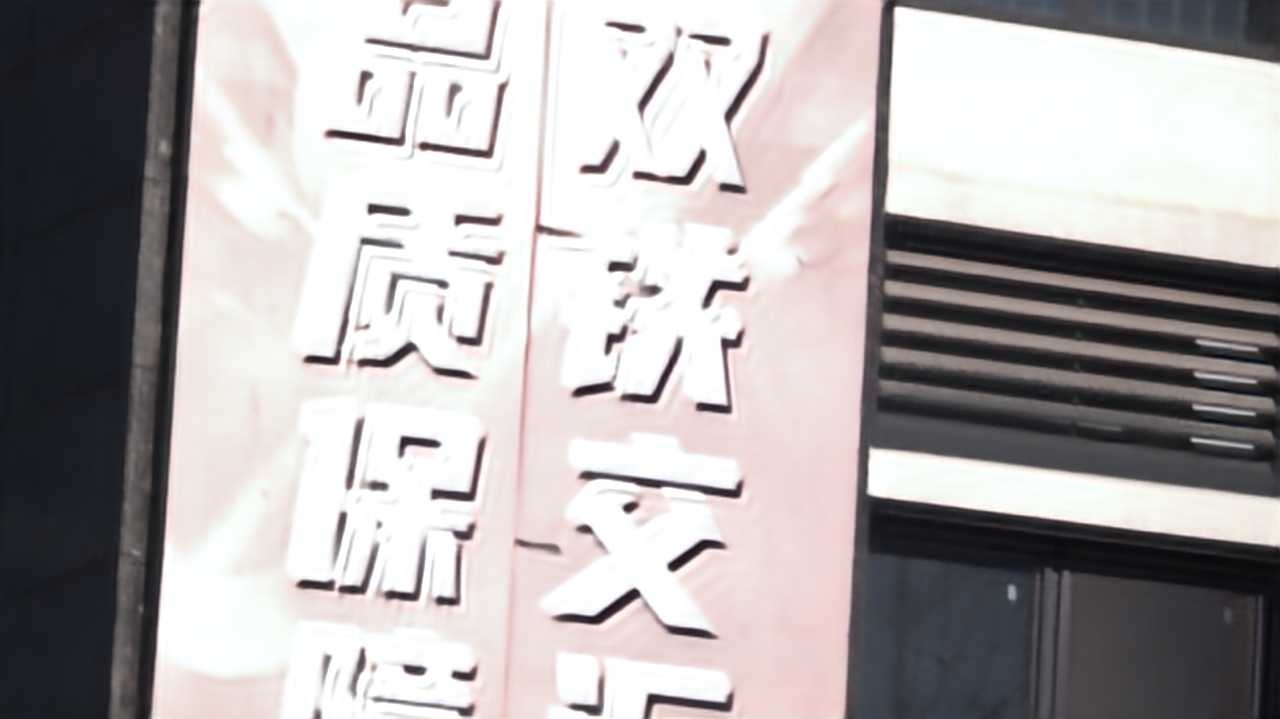}\\[2pt]{\scriptsize Ours}
\end{minipage}\hfill
\begin{minipage}{0.116\textwidth}\centering
\includegraphics[width=\textwidth]{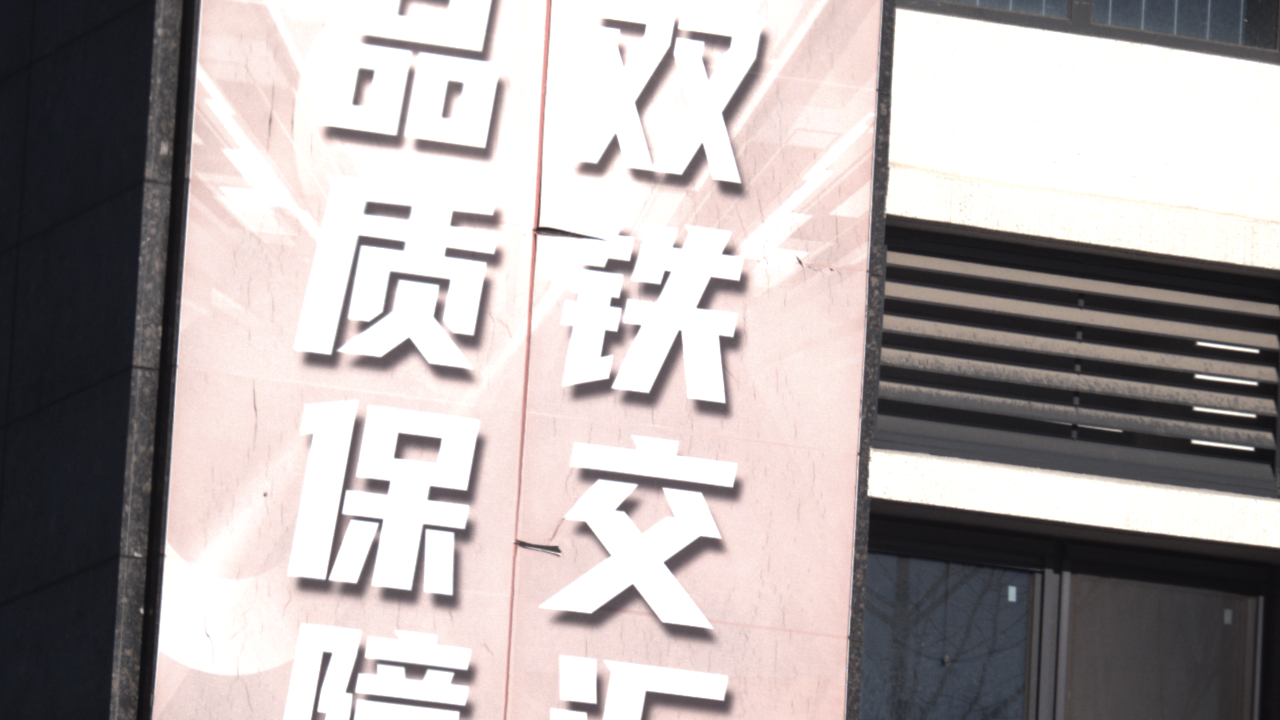}\\[2pt]{\scriptsize GT}
\end{minipage}
\caption{\Add{Qualitative comparison on EventAid-B. Blind methods (NAFNet, Restormer) leave residual blur, EFNet introduces over-sharpening and brightness shifts, and our method restores sharp detail while preserving the original tone.}}
\label{fig:deblur_qual_eventaid}
\vspace{-17pt}
\end{figure*}

\Add{Table~\ref{table:deblur} reports results on EventAid-B. Image-only methods cannot recover severely blurred regions and trail the event-assisted methods in PSNR. EFNet is sensitive to the mismatch between its training event statistics and EventAid-B's degraded events: its PSNR falls below the parameter-tuned EDI baseline, though fine-tuning on REBlur raises its SSIM above every other method, including ours. With only 0.6M parameters, our method attains the highest PSNR, attributable to the ODF-estimated kernel's accuracy rather than network capacity. At matched capacity, pairing the same estimated kernel with the plain USRNet-tiny backbone reaches only 26.35/0.843, so our event-mask conditioning still contributes beyond capacity or kernel accuracy alone. Our SSIM trails EFNet-REBlur, plausibly because that fine-tuning targets REBlur's specific structural statistics, a trade-off between a more accurate motion-derived kernel and a dataset-tuned prior (qualitative comparison in Fig.~\ref{fig:deblur_qual_eventaid}).}

\begin{table}[!t]
\caption{{Comparison of single image motion deblurring methods on blurry images with event data and ground-truth blur kernels. \Add{Bold marks the best result per column (gt kernel rows are an oracle upper bound).}}}
\vspace{-11pt}
\centering
\small
\resizebox{0.48\textwidth}{!}{
\setlength{\tabcolsep}{10pt}
\renewcommand\arraystretch{1.0}
\begin{tabular}{ l c  c  c  c }
\bottomrule[0.15em]
\rowcolor{tableHeadGray}
\textbf{Method} &\textbf{Events} & \textbf{PSNR} $\uparrow$ & \textbf{SSIM} $\uparrow$ & \textbf{\#Param} \\ \hline
EDI~\cite{edi_pan}                  &\cmark & 30.54 & 0.856 & - \\
Restormer~\cite{Zamir2021Restormer}          & \xmark         & 30.34  & 0.861 & 26.1M \\
NAFNet-32~\cite{chen2022simple}                 & \xmark         & 29.93  & 0.848 &  17.1M \\
EFNet~\cite{sun2022event}                    & \cmark         & 31.67 &  0.854 & 8.5M \\
ODF kernel + DWDN~\cite{dong2021dwdn} & \cmark         & 31.79  & 0.874  & 7.0M \\
ODF kernel + RAM~\cite{terris2025ram} & \cmark         & 33.00  & 0.897  & 35.6M \\
\Add{\textbf{Ours}} (ODF kernel + event mask)   & \cmark         & 32.33  & 0.885  & 0.6M \\
DWDN w/ gt kernel~\cite{dong2021dwdn} & \cmark         & 32.58  & 0.890  & 7.0M \\
RAM w/ gt kernel~\cite{terris2025ram} & \cmark         & \Add{\textbf{33.39}}  & \Add{\textbf{0.903}}  & 35.6M \\
Ours w/ gt kernel   & \cmark         & 32.92  & 0.895  & 0.6M \\
\hline

\end{tabular}
}
\vspace{-15pt}
\label{table:deblur collected}
\end{table}

\begin{figure*}[!t]
\centering
\begin{minipage}{0.19\textwidth}\centering
\includegraphics[width=\textwidth]{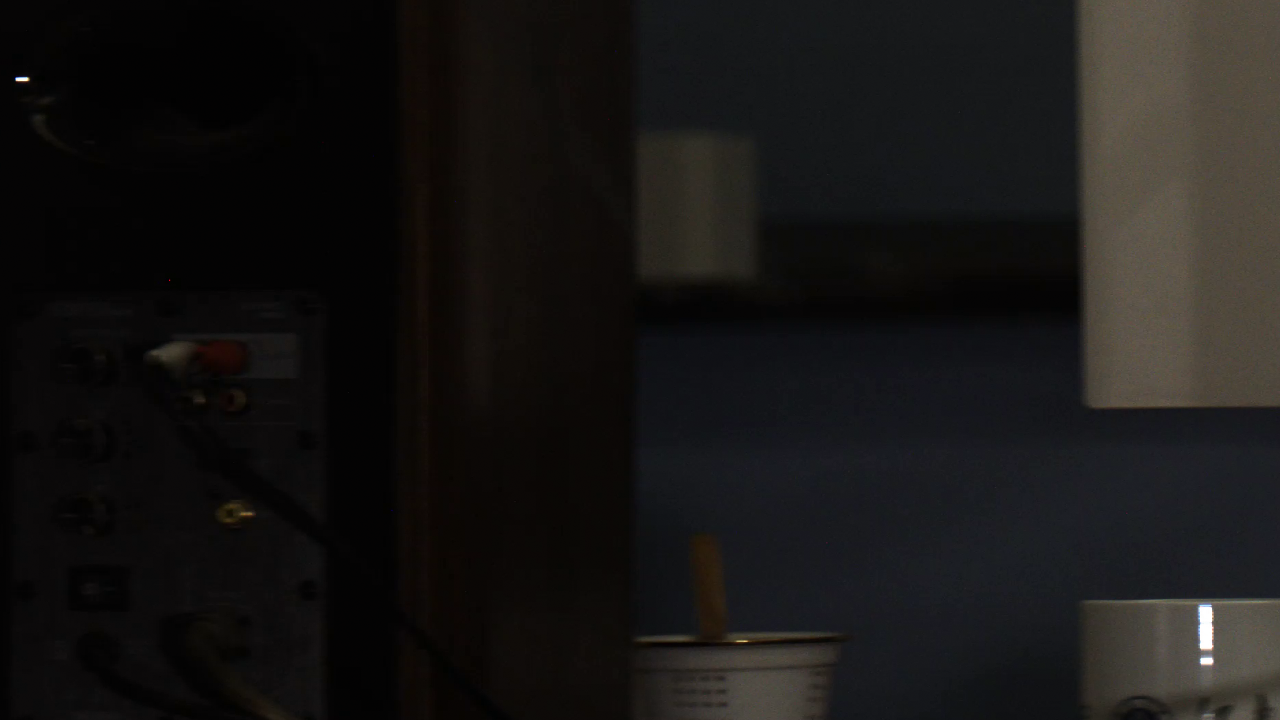}\\[2pt]{\scriptsize Blurred}
\end{minipage}\hfill
\begin{minipage}{0.19\textwidth}\centering
\includegraphics[width=\textwidth]{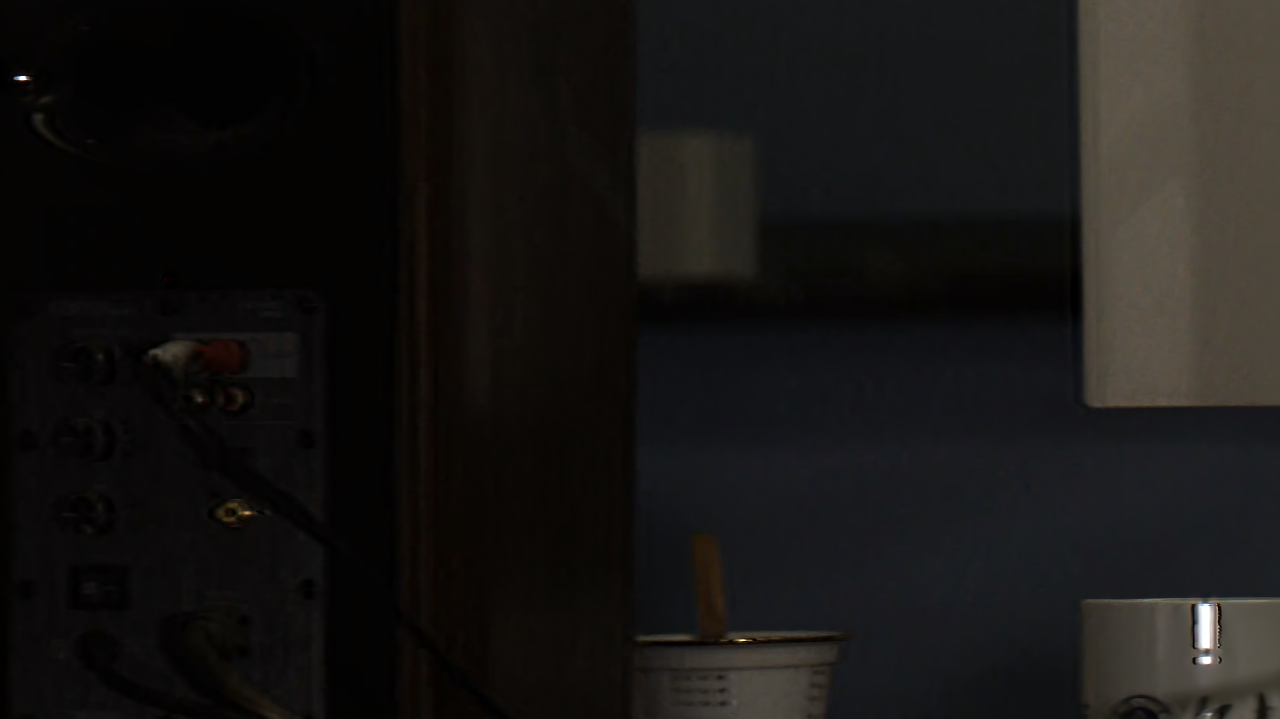}\\[2pt]{\scriptsize ODF kernel + DWDN}
\end{minipage}\hfill
\begin{minipage}{0.19\textwidth}\centering
\includegraphics[width=\textwidth]{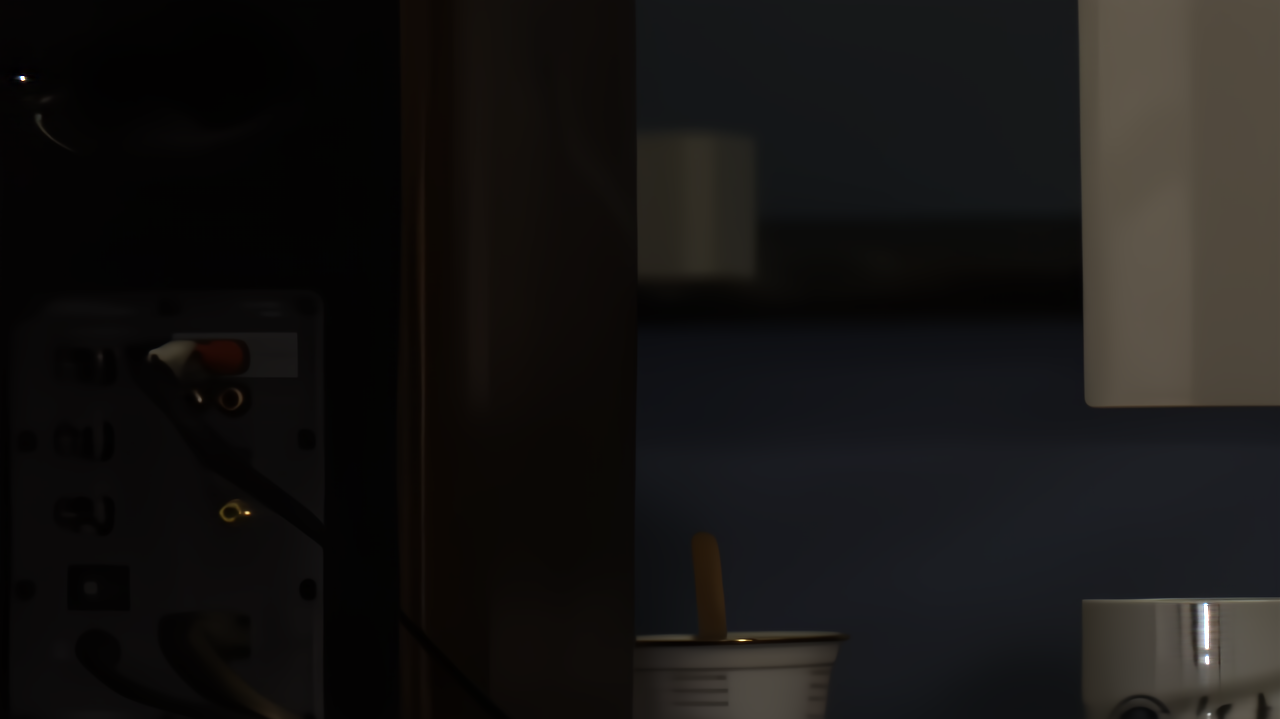}\\[2pt]{\scriptsize ODF kernel + RAM}
\end{minipage}\hfill
\begin{minipage}{0.19\textwidth}\centering
\includegraphics[width=\textwidth]{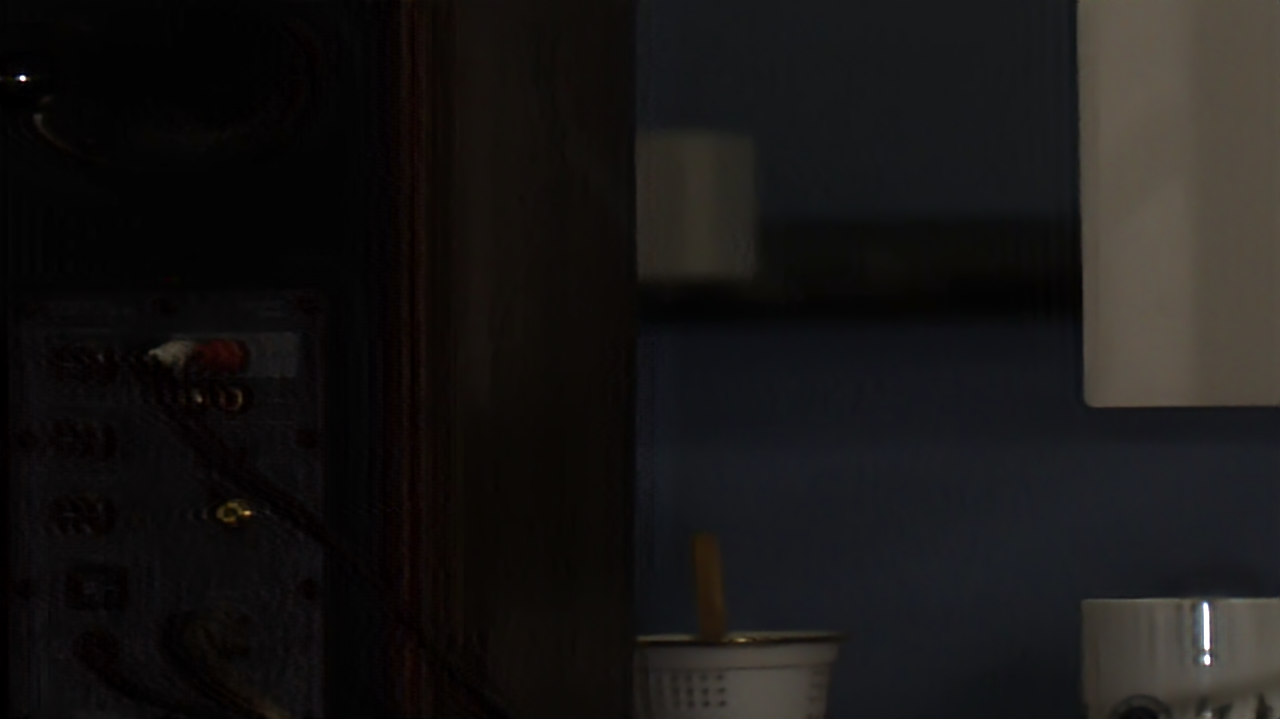}\\[2pt]{\scriptsize Ours}
\end{minipage}\hfill
\begin{minipage}{0.19\textwidth}\centering
\includegraphics[width=\textwidth]{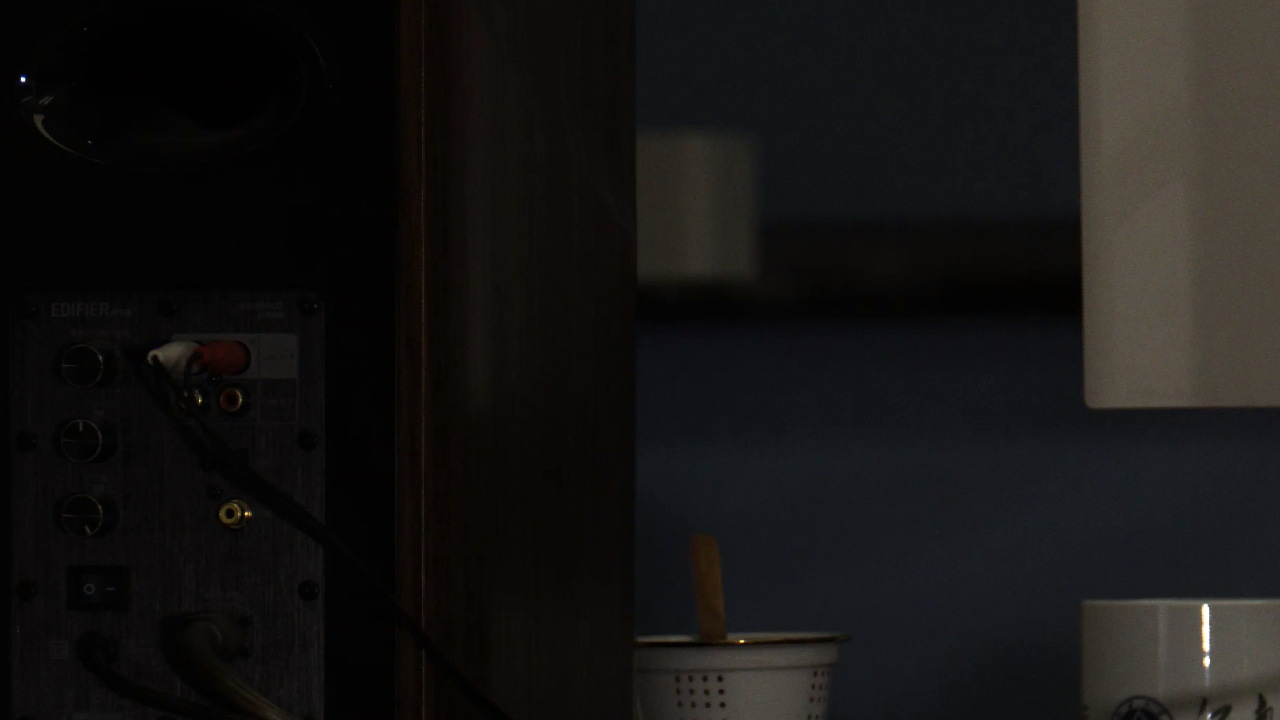}\\[2pt]{\scriptsize GT}
\end{minipage}
\caption{\Add{Qualitative comparison on our real captured dataset, using our estimated blur kernel. Our lightweight network controls edge artifacts better than RAM, though RAM recovers marginally sharper texture.}}
\label{fig:deblur_qual_real}
\vspace{-17pt}
\end{figure*}

\Add{Table~\ref{table:deblur collected} evaluates the same networks on our real dataset, with both our estimated kernel and the ground-truth trajectory. EDI's closed-form reconstruction attains the second-highest PSNR among the baselines (30.54, behind EFNet's 31.67) and a competitive SSIM, yet still trails every ODF-kernel-based method. Replacing the ground-truth kernel with our estimated one costs each network under 1~dB in PSNR and 0.02 in SSIM, confirming ODF motion estimation recovers kernels close to the true motion. Our network trails the much larger RAM by under 1~dB in PSNR at both kernel settings, though it outperforms DWDN despite an order of magnitude fewer parameters than either, a capacity trade-off rather than a kernel-quality one: both receive the identical kernel, so the gap traces to RAM's 35.6M parameters and plug-and-play restoration affording more capacity than our 0.6M network spends (qualitative comparison in Fig.~\ref{fig:deblur_qual_real}).}

\noindent\textbf{Speed.}
\Add{ODF motion estimation processes each event in under 0.7~\textmu s on a single CPU thread (14th Gen Intel Core i7-14650HX), since it only looks up and averages precomputed distance vectors instead of solving an optimization problem. The deblurring network runs at 55~ms per $1280\times720$ image on an Nvidia A800 GPU and reaches real time, above 60~fps, at $512\times512$.}

\Add{We do not report a separate ablation study: ODF motion estimation follows from a closed-form derivation rather than empirically tuned modules, and the deblurring network's two main design choices are already isolated within the main comparison tables. Training on motion-estimation-noise kernels is isolated by the estimated- and ground-truth-kernel rows in Table~\ref{table:deblur collected}, which differ by under 1~dB in PSNR and 0.02 in SSIM; the event-mask conditioning is isolated by the plain-USRNet-tiny row in Table~\ref{table:deblur}, which trails our network by 0.88~dB PSNR and 0.017 SSIM at matched parameter count.}

\subsection{Pupil Motion Estimation for Eye Tracking}
\begin{figure}[!t]
    \centering
    \begin{minipage}{0.155\textwidth}\centering
    \includegraphics[width=\textwidth]{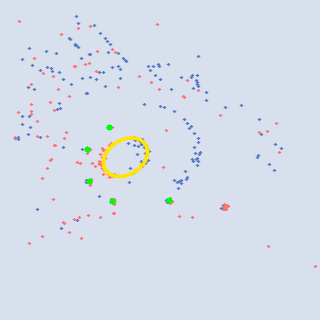}\\[2pt]{\scriptsize $t_1$}
    \end{minipage}\hspace{1mm}
    \begin{minipage}{0.155\textwidth}\centering
    \includegraphics[width=\textwidth]{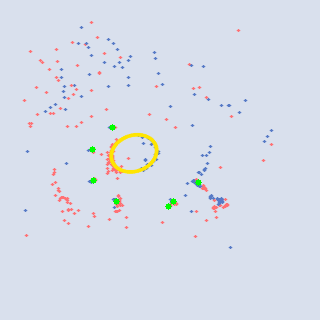}\\[2pt]{\scriptsize $t_2$}
    \end{minipage}\hspace{1mm}
    \begin{minipage}{0.155\textwidth}\centering
    \includegraphics[width=\textwidth]{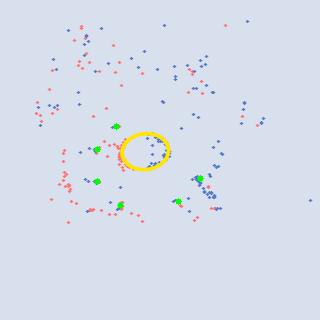}\\[2pt]{\scriptsize $t_3$}
    \end{minipage}
    \caption{\Add{Pupil and glint tracking on near-eye event data at three time steps $t_1$--$t_3$. Events are shown in red (positive) and blue (negative), with the fitted pupil ellipse in yellow and located glints in green.}}
    \label{fig:eye_tracking_results}
    \vspace{-17pt}
\end{figure}

\Add{We evaluate pupil tracking accuracy on the public near-eye event dataset of Angelopoulos et al.~\cite{angelopoulos2021event} (24 subjects), using EllSeg~\cite{kothari2021ellseg} pseudo-labels on the frame images as reference, since the original dataset's per-frame pupil extraction is not public. Tracking is initialized from the first valid ellipse after each blink and evaluated against the temporally nearest EllSeg label. Across subjects, the median pupil IoU is 0.85 (left) and 0.86 (right), with a median center error of 1.45 and 1.41~pixels, and the longest continuous track reaches 1745 frames (about 70~s at 25~fps), showing tracking sustained through long fixations. On a near-eye AR prototype using a Prophesee GenX320 event sensor~\cite{prophesee2023genx320}, initialized from an E2VID~\cite{rebecq2021e2vid} reconstruction, the tracker recovers both pupil and glint positions throughout the recorded sequences (Fig.~\ref{fig:eye_tracking_results}), processing 28~\textmu s per event on a single thread, dropping to 10~\textmu s on separate threads, comfortably real time. We further compare the power draw of a near-eye module built around the GenX320 against one around an OmniVision OVM6211 frame sensor, with the frame sensor's LED lit only during its own exposure window (Table~\ref{table:power}). The event sensor draws under a third of the frame sensor's power at every rate tested, but its continuously-lit LED draws more than the frame sensor's duty-cycled LED at 60~fps, since our glint detection is not yet optimized for duty cycling. The sensor-side saving still dominates, so the event-driven module draws less overall, 127.1~mW against 215.0~mW at 60~fps, widening as the frame rate increases: the tracker's low data rate lowers system power even though it does not win on every sub-component.}

\begin{table}[!t]
\caption{\Add{Hardware power comparison of a frame-camera and an event-camera near-eye tracking module (bold: lowest total power).}}
\vspace{-11pt}
\centering
\small
\resizebox{0.48\textwidth}{!}{
\setlength{\tabcolsep}{5pt}
\renewcommand\arraystretch{1.0}
\begin{tabular}{l c c c c c}
\bottomrule[0.15em]
\rowcolor{tableHeadGray}
\textbf{Sensor} & \textbf{Rate (fps)} & \textbf{Sensor $\downarrow$ (mW)} & \textbf{LED duty} & \textbf{LED $\downarrow$ (mW)} & \textbf{Total $\downarrow$ (mW)} \\ \hline
\multirow{3}{*}{Frame camera} & 60  & 161.8 & 30\% & 53.2 & 215.0 \\
 & 90  & 163.6 & 45\% & 75.6 & 239.2 \\
 & 120 & 165.4 & 60\% & 100.8 & 266.2 \\ \hline
Event camera & -- & 51.2 & 100\% & 75.9 & \textbf{127.1} \\ \hline
\end{tabular}
}
\vspace{-15pt}
\label{table:power}
\end{table}

\section{Conclusion}
\label{sec:conclusion}

\Add{We presented ODF motion estimation, which replaces iterative optimization with a single averaging step over a precomputed distance vector field, and showed this core module generalizes to two downstream tasks. For deblurring, the recovered trajectory yields a blur kernel for a lightweight network trained on motion-estimation-noise simulation, matching or exceeding prior event-assisted methods at a fraction of their parameter count. For eye tracking, the same distance vectors filter events directionally for stable, low-power pupil and glint tracking. Motion estimation itself runs at sub-microsecond latency per event, so the low-latency advantage of event cameras is preserved end to end rather than traded away for accuracy.}

\noindent\textbf{Limitations.}
\Add{Our method assumes globally consistent motion, sufficient orientation diversity among tracked edges, and enough events to constrain the estimate, so dynamic scenes, templates dominated by one edge orientation, and sparse event regions each degrade accuracy. Both applications also assume 2-DoF translation rather than general 6-DoF motion, valid for gimbal- or PTZ-mounted cameras and similar platforms but not for larger rotation or affine motion, which we leave to future work.}

\ifCLASSOPTIONcaptionsoff
  \newpage
\fi

{\scriptsize
\bibliographystyle{ieee_fullname}
\bibliography{main}
}

\end{document}